\documentclass[sigconf,preprint]{acmart}
\AtBeginDocument{%
  }

\setcopyright{acmlicensed}
\copyrightyear{2018}
\acmYear{2018}
\acmDOI{XXXXXXX.XXXXXXX}
\acmConference[Conference acronym 'XX]{Make sure to enter the correct
  conference title from your rights confirmation email}{June 03--05,
  2018}{Woodstock, NY}
\acmISBN{978-1-4503-XXXX-X/2018/06}

\usepackage{tabularx} 
\usepackage{amsmath}        
\usepackage{multirow}
\usepackage{multicol}
\usepackage{subcaption}
\usepackage{cuted} 
\usepackage{makecell}

\usepackage[most]{tcolorbox} 
\tcbuselibrary{listingsutf8}

\usepackage{pifont}  
\newcommand{\xmark}{\ding{55}} 
\newcommand{\cmark}{\ding{51}} 

\usepackage[ruled,vlined]{algorithm2e} 
\newtcolorbox{textbox}{
  enhanced,
  colback=gray!5,
  colframe=gray!50,
  boxrule=0.5pt,
  arc=4pt,
  auto outer arc,
  fontupper=\scriptsize
}

\newtcolorbox{tablebox}[2]{
  enhanced,
  colback=gray!5,
  colframe=gray!50,
  boxrule=0.5pt,
  arc=4pt,
  auto outer arc,
  fontupper=\scriptsize,
  listing only,
  title=#2,
  title style={fontupper=\bfseries\scriptsize}, 
  listing options={
    language=#1,
    basicstyle=\scriptsize\ttfamily,
    breaklines=true,
    columns=fullflexible
  }
}
\usepackage{gradient-text}
\newcommand{\fintagging}{\gradientRGB{FinTagging}{57,126,255}{240,89,215}\xspace}

\begin{document}

\title{\textsc{\fintagging}: Benchmarking LLMs for Extracting and Structuring Financial Information}

\author{Yan Wang}
\affiliation{%
  \institution{\normalsize The Fin AI}
  \country{\normalsize USA}
}
\email{wy2266336@gmail.com}

\author{Lingfei Qian}
\authornote{Corresponding authors.}
\affiliation{%
  \institution{\normalsize The Fin AI}
  \country{\normalsize USA}
}
\email{lfqian94@gmail.com}

\author{Xueqing Peng}
\author{Yang Ren}
\affiliation{%
  \institution{\normalsize The Fin AI}
  \country{\normalsize USA}
}

\author{Yi Han}
\affiliation{%
  \institution{\normalsize Georgia Institute of Technology}
  \country{\normalsize USA}}
  
\author{Keyi Wang}
\affiliation{%
  \institution{\normalsize Columbia University}
  \country{\normalsize USA}}

\author{Dongji Feng}
\affiliation{%
  \institution{\normalsize California State University}
  \country{\normalsize USA}
}

\author{Fengran Mo}
\affiliation{%
  \institution{\normalsize University of Montreal}
  \country{\normalsize Canada}}

\author{Shengyuan Lin}
\affiliation{%
 \institution{\normalsize Carnegie Mellon University}
 \country{\normalsize USA}}

\author{Qinchuan Zhang}
\affiliation{%
  \institution{\normalsize Rensselaer Polytechnic Institute}
  \country{\normalsize USA}}

\author{Kaiwen He}
\author{Chenri Luo}
\author{Jianxing Chen}
\author{Junwei Wu}
\affiliation{%
  \institution{\normalsize Columbia University}
  \country{\normalsize USA}}


\author{Chen Xu}
\author{Ziyang Xu}
\affiliation{%
  \institution{\normalsize The Fin AI}
  \country{\normalsize USA}
}

\author{Jimin Huang}
\affiliation{%
   \institution{The University of Manchester}
   \city{Manchester}
   \country{United Kingdom}
}
\affiliation{%
  \institution{\normalsize The Fin AI}
  \country{\normalsize USA}
}

\author{Guojun Xiong}
\affiliation{%
  \institution{\normalsize Harvard University}
  \country{\normalsize USA}}

\author{Xiao-Yang Liu}
\affiliation{%
  \institution{\normalsize Columbia University}
  \country{\normalsize USA}}

\author{Qianqian Xie}
\affiliation{%
  \institution{\normalsize The Fin AI}
  \country{\normalsize USA}
}

\author{Jian-Yun Nie}
\affiliation{%
  \institution{\normalsize University of Montreal}
  \country{\normalsize Canada}}

\renewcommand{\shortauthors}{Yan et al.}

\begin{abstract}

Accurate interpretation of numerical data in financial reports is critical for markets and regulators. Although XBRL (eXtensible Business Reporting Language) provides a standard for tagging financial figures, mapping thousands of facts to over 10k US-GAAP concepts remains costly and error-prone. Existing benchmarks oversimplify this task as flat, single-step classification over small subsets of concepts, ignoring the hierarchical semantics of the taxonomy and the structured nature of financial documents. Consequently, these benchmarks fail to evaluate Large Language Models (LLMs) under realistic reporting conditions. To bridge this gap, we introduce \textsc{{\fintagging}}, the first comprehensive benchmark for structure-aware and full-scope XBRL tagging. We decompose the complex tagging process into two subtasks: (1) \textbf{FinNI} (Financial Numeric Identification), which extracts entities and types from heterogeneous contexts (text and tables); and (2) \textbf{FinCL} (Financial Concept Linking), which maps extracted entities to the full US-GAAP taxonomy. This two-stage formulation enables a fair assessment of LLMs' capabilities in numerical reasoning and taxonomy alignment. Evaluating diverse LLMs in zero-shot settings reveals that while models generalize well in extraction, they struggle significantly with fine-grained concept linking, highlighting critical limitations in domain-specific structure-aware reasoning. Code is available at GitHub~\footnote{\url{https://github.com/The-FinAI/FinTagging}} and datasets are available at Hugging Face~\footnote{\url{https://huggingface.co/collections/TheFinAI/fintagging}}.

\end{abstract}

\ccsdesc[500]{Computing methodologies~Natural language processing}
\ccsdesc[500]{Computing methodologies~Information extraction}
\ccsdesc[500]{Information systems~Information extraction}
\ccsdesc[500]{Information systems~Top-k retrieval in databases}

\keywords{XBRL tagging, Benchmark, Large language model, Information extraction, Information retrieval, Reranking}


\maketitle

\section{Introduction}
\label{intro}
Although LLMs excel at question answering~\cite{qian2025fino1}, information extraction~\cite{balasubramanian2025leveraging}, and long-document summarization~\cite{yuan2025strucsum}, their ability to perform financial tagging remains underexplored. Unlike open-ended generation, tagging requires committing each reported number to a precise semantic interpretation, grounding it in mixed table and text evidence, and aligning it to standardized concepts interpretable by downstream systems.

\begin{figure*}[h]
  \centering
  \setlength{\abovecaptionskip}{2pt}
  \setlength{\belowcaptionskip}{0pt}
  \includegraphics[width=0.85\linewidth]{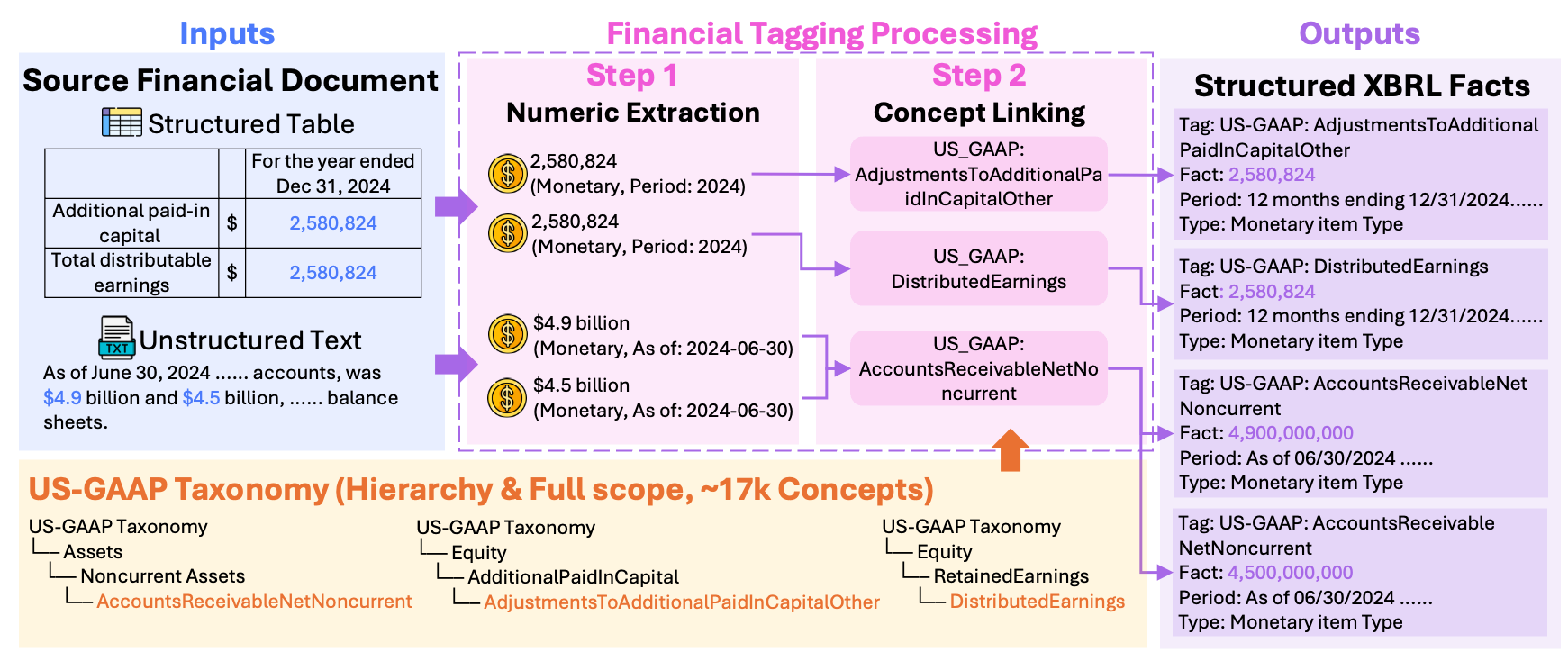}
  \caption{An example of Financial Tagging in realistic scenarios, where the blue numbers mark facts to be tagged.}
  \label{fig:example-figure}
\end{figure*}

Each year, over 2 million companies publish financial reports disclosing earnings, expenses, and liabilities. However, inconsistent terminology complicates reliable interpretation. XBRL, introduced in 1999, defines the global standard for machine-readable financial reporting~\cite{richards2006introduction} and is now adopted in 65 countries\footnote{\url{https://www.xbrl.org/the-standard/what/what-is-xbrl/}}
. In practice, XBRL tagging requires grounding thousands of reported numbers per filing in mixed evidence and aligning them to over 17,000 standardized taxonomy concepts (Figure~\ref{fig:example-figure}), a process that remains largely manual and error-prone\footnote{\url{https://xbrl.us/wp-content/uploads/2023/03/DQC-SECMeetingNotes-20240314.pdf}}.

Nevertheless, existing XBRL tagging benchmarks emphasize annotation convenience over semantic difficulty. Many reduce tagging to classification over truncated taxonomies and remove tabular structure that carries critical evidence in real filings (Table~\ref{tab:benchmark_comparison}). As a result, evaluation is limited to a small subset of common concepts, obscuring long-tailed generalization.
Moreover, prior work typically formulates tagging as closed-set extreme classification~\cite{loukas2022finer,sharma2023financial}, which becomes brittle when the full taxonomy cannot be represented in-context and increases the risk of invalid or hallucinated tags. Our ablations confirm this limitation: even strong LLMs show near-zero performance under full-taxonomy settings. In addition, most benchmarks omit tables entirely, despite tables being the primary carrier of numerical disclosures, and thus fail to evaluate the joint reasoning over structured tables and narrative context required in realistic tagging~\cite{chen2021finqa,zhu2021tat}.

\begin{table*}[h]
\setlength{\abovecaptionskip}{3pt}  
\centering
\caption{Detailed comparison of financial NLP benchmarks across task types, sources, and structural capabilities. ``Num.Reasoning'' indicates the numerical reasoning. ``Struct.IE'' denotes the structured information extraction.}
\label{tab:benchmark_comparison}
\resizebox{.85\linewidth}{!}{\begin{tabular}{lllllllccc}
\toprule
\textbf{Benchmark} & \textbf{Scenario} & \textbf{Data Source} & \textbf{Task} & \textbf{Modality} & 
\makecell{\textbf{\#Entity}\\\textbf{Label}} & \makecell{\textbf{\#Taxonomy}\\\textbf{Label}} & \makecell{\textbf{Num.}\\\textbf{Reasoning?}} & 
\makecell{\textbf{Struct.}\\\textbf{IE?}} & \makecell{\textbf{Concept}\\\textbf{Linking?}} \\
\midrule
FinQA~\cite{chen2021finqa}             & Decision making           & SEC 10-K                     & QA                           & text/table  & 0  & 0      & \cmark & \xmark & \xmark \\
ConvFinQA~\cite{chen2022convfinqa}         & Decision making           & SEC 10-K                     & QA                           & text/table  & 0  & 0      & \cmark & \xmark & \xmark \\
TAT-QA~\cite{zhu2021tat}            & Financial analysis        & Chinese financial reports    & QA                           & text/table  & 0  & 0      & \cmark & \xmark & \xmark \\
DocVQA~\cite{mathew2021docvqa}            & Enterprise automation     & Financial document images    & QA                           & text/image  & 0  & 0      & \cmark & \xmark & \xmark \\
FiNER-ORD~\cite{shah2023finer}         & Financial tagging         & Chinese financial disclosures& Classification               & text        & 49 & 0      & \xmark & \xmark & \xmark \\
FinRED~\cite{sharma2022finred}            & Knowledge graph construction & English financial news   & Classification               & text        & 38 & 0      & \xmark & \xmark & \xmark \\
FiNER~\cite{loukas2022finer}         & XBRL tagging              & SEC 10-K                     & Extreme classification               & text        & 0  & 139 / 17,688    & \cmark & \xmark & \xmark \\
FNXL~\cite{sharma2023financial}              & XBRL tagging              & SEC 10-K                     & Extreme classification               & text        & 0  & 2,800 / 17,688   & \cmark & \xmark & \xmark \\
FINTAGGING & Financial \& XBRL tagging & SEC 10-K                   & IE + Alignment & text/table  & 5  & 17,688 / 17,688  & \cmark & \cmark & \cmark \\
\bottomrule
\end{tabular}}
\end{table*}

To address these issues, we introduce \textsc{\fintagging}, the first LLM-oriented benchmark that evaluates end-to-end financial tagging over mixed table and text inputs with alignment to the full US-GAAP taxonomy. Unlike prior work~\cite{loukas2022finer,sharma2023financial}, \textsc{\fintagging} reframes tagging as an \emph{extract-and-align} process rather than flat classification, enabling evaluation across all 17k+ concepts.
\textsc{\fintagging} adopts a two-stage formulation. \textbf{Financial Numerical Identification (FinNI)} focuses on extracting numerical facts with supporting evidence, while \textbf{Financial Concept Linking (FinCL)} maps each fact to the correct US-GAAP concept via semantic alignment with taxonomy definitions. We release two expert-verified evaluation sets, \textit{FinNI-eval} and \textit{FinCL-eval}, derived from real-world XBRL.


We conclude our main contributions as follows:
(1) \textbf{A New Formulation:} We introduce \textsc{\fintagging}, the first benchmark that reframes XBRL tagging into a \textit{two-stage ``Extract-and-Align'' paradigm}, enabling full-scope evaluation against 17k+ US-GAAP concepts, a scale previously intractable for zero-shot LLMs.
(2) \textbf{High-Fidelity Resources:} We release two expert-verified datasets, \textit{FinNI-eval} and \textit{FinCL-eval}, which incorporate complex tabular structures to assess joint reasoning across structured and unstructured financial modalities.
(3) \textbf{Diagnostic Insights:} By benchmarking 13 state-of-the-art LLMs, we identify a critical \textbf{``knowledge-alignment gap''}, demonstrating that while LLMs excel at numeric extraction, they require our alignment framework to prevent performance collapse during large-scale taxonomy mapping.

\section{Related Work}
\label{related}

Prior XBRL tagging research primarily formulates the task as extreme multi-class classification over truncated label sets, as in FiNER and FNXL~\cite{loukas2022finer,sharma2023financial}, with methods such as GalaXC, SECBERT, and AttentionXML~\cite{saini2021galaxc,loukas2022finer,sharma2023financial} addressing label sparsity via hierarchical modeling. However, these closed-set formulations do not reflect the semantic reasoning required to navigate the full 17k+ US-GAAP taxonomy and are poorly aligned with LLM capabilities. In parallel, financial document understanding has advanced from entity and relation extraction~\cite{shah2023finer,sharma2022finred} to joint text–table numerical reasoning in QA benchmarks such as FinQA, TAT-QA, and DocVQA~\cite{chen2021finqa,zhu2021tat,mathew2021docvqa}, yet these tasks do not require alignment to standardized accounting concepts. Recent LLM-based tools such as XBRL-Agent~\cite{han2024xbrlagent} explore filing analysis but stop short of end-to-end fact-to-taxonomy evaluation. In contrast, \textsc{\fintagging} reframes XBRL tagging as a two-stage extract-and-align task, explicitly targeting multi-modal fact extraction and fine-grained alignment to official US-GAAP definitions.

\section{\textsc{\fintagging}}
\label{ben}

\subsection{Task Formulation}
\label{formulation}

Formally, given a financial document $D$ containing text and structured tables, a set of predefined XBRL value types $L$, and a taxonomy $\mathcal{T}$ of financial concepts, the XBRL tagging task is to identify all numerical values in $D$ and annotate each with a structured triplet $\{ \texttt{Fact}, \texttt{Type}, \texttt{Tag}\}$. 
Here, \texttt{Fact} denotes the extracted numerical value as expressed in context, \texttt{Type} specifies its XBRL data type in $L$, and \texttt{Tag} refers to the linked concept in $\mathcal{T}$ (e.g., \texttt{us-gaap:CashAndDueFromBanks}). 
Each numerical fact is assigned a single, most specific taxonomy concept.
The value type set $L$ includes \textit{monetaryItemType}, \textit{percentItemType}, \textit{sharesItemType}, \textit{perShareItemType}, and \textit{integerItemType}, corresponding to monetary amounts, ratios, share counts, per-share values, and integer quantities, respectively.
We formalize the task as a mapping:

\begin{equation}
    f: (D, L, \mathcal{T}) \mapsto \left\{ (\texttt{Fact}_i, \texttt{Type}_i, \texttt{Tag}_i) \right\}_{i=1}^{n}
\end{equation}
where each triplet corresponds to a financial value mention extracted from $D$ and semantically grounded in $\mathcal{T}$.

 Inspired by information extraction and alignment works~\cite{wu2019scalable,wadden2019entity,mukherjee2022ectsum,wang2022conditional} and after discussing with financial reporting specialists, we formulated \textsc{\fintagging} into two sub-tasks: \textbf{Financial Numeric Identification (FinNI)}, a multi-modal numerical information extraction task, detects numerical value in $D$ and classifies each with its appropriate \texttt{Type}. \textbf{Financial Concept Linking (FinCL)}, a numerical entity normalization task, then associates each identified value with its most appropriate \texttt{Tag} in $\mathcal{T}$ based on contextual and structural cues.

\subsubsection{Financial Numeric Identification (FinNI)}
The first subtask of \textsc{\fintagging} focuses on identifying numerical values in a financial document and assigning each a coarse-grained value data type. This corresponds to detecting the \texttt{Fact} and \texttt{Type} components of each triplet $\{\texttt{Fact}, \texttt{Type},\texttt{Tag}\}$ defined in the overall task. We formalize this subtask as a mapping:

\begin{equation}
    f_{\text{FinNI}}: (D = (S, T),\ L) \mapsto \left\{ (e_i, l_i) \right\}_{i=1}^{k}
\end{equation}
where $S$ and $T$ represent the textual and tabular components of the document, respectively, and $L$ is the set of predefined value data types. Each $e_i$ is a numerical entity extracted from either $S$ or $T$, and $l_i \in L$ denotes its assigned data type.

\subsubsection{Financial Concept Linking (FinCL)}
\label{fincl}
Building on the output of the {FinNI} subtask, the goal of {FinCL} is to semantically ground each identified numerical entity $e$ by linking it to a concept $\hat{c}$ in a predefined financial taxonomy, which is the \texttt{Tag} component of each triplet $\{\texttt{Fact}, \texttt{Type},\texttt{Tag}\}$ defined in the overall task. Formally, we define the mapping as:
\begin{equation}
    f_{\text{FinCL}}: (e, l, C_e, \mathcal{T}) \mapsto \hat{c}
\end{equation}
where $e$ is a numerical entity extracted from $D = (S, T)$, and $l \in L$ is its associated data type provided by the {FinNI} stage. 
$C_e$ represents the \textbf{structural context} of $e$, which encompasses linguistic dependencies for narrative mentions and hierarchical row-column coordinates for tabular mentions. 
$\mathcal{T} = \{c_1, c_2, \dots, c_n\}$ denotes the full-scope US-GAAP taxonomy containing $n$ uniquely defined and semantically grounded concepts. 
The objective of {FinCL} is to identify the optimal concept $\hat{c} \in \mathcal{T}$ that maximizes the semantic alignment between the fact $e$ and its specific context $C_e$ within the high-dimensional taxonomy space.

\subsection{Raw Data Collection}
\label{collection}

We compiled 142 annual 10-K filings from publicly listed U.S. companies submitted to the U.S. Securities and Exchange Commission (SEC)\footnote{\url{https://www.sec.gov/}} in 2023–2024 (Table~\ref{tab:statistic_finRep}). The collection covers all 11 major industry sectors, with companies primarily based in the United States and a small portion internationally. Parsing the filings with BeautifulSoup yields 319,893 narrative sentences and 21,576 financial tables.
All filings follow the SEC Inline XBRL (iXBRL) format, which embeds machine-readable tags within HTML disclosures and links textual and tabular values to US-GAAP taxonomy concepts. However, these regulatory tags are designed for compliance rather than model evaluation, providing limited negative coverage and substantial structural noise.
We therefore repurpose the raw iXBRL metadata to construct a standardized, high-fidelity evaluation suite for financial fact extraction and concept linking.



\begin{table}[h]
\setlength{\abovecaptionskip}{3pt}  
\footnotesize
\caption{Statistics of collected financial reports.}
\label{tab:statistic_finRep}
\centering
\begin{tabular}{lc}
\toprule
\textbf{Item} & \textbf{Value} \\
\midrule
Report type   & 10-K \\
Period        & 2023-02-13 to 2025-02-13 \\
\#Company     & 142 \\
\#Covered sector & 11 \\
\#Covered jurisdiction & 31 States + outside US \\
\#Sentence    & 319,893 \\
\#Char        & 75,748,949 \\
\#Table       & 21,576 \\
\bottomrule
\end{tabular}
\end{table}

\subsection{Large-Scale Automated Annotation}
\label{filtered_data}

Our automated pipeline produces a large, high-fidelity corpus reflecting real financial disclosures. As summarized in Table~\ref{tab:overall_benchmark_stats}, the dataset contains 15,986 narrative sentences and 12,801 tables, yielding 261,457 numerical entities linked to 3,953 unique US-GAAP concepts. Including both valid facts and naturally occurring noise, it supports a realistic evaluation of LLMs under regulatory-style conditions.

\begin{table}[h]
\scriptsize
\caption{Statistical information for the original dataset in our benchmark. 
Tokens are calculated using the ``cl100k\_base'' tokenizer ($\pm$ standard deviation).}
\label{tab:overall_benchmark_stats}
\centering
\resizebox{0.8\linewidth}{!}{
\begin{tabular}{lcc}
\toprule
\textbf{Item} & \textbf{Sentence} & \textbf{Table} \\
\midrule
Positive instances     & 7,768 & 8,709 \\
Negative instances     & 8,218 & 4,092 \\
Avg. Tokens/S          & $80.91 \pm 63.62$ & $1212.42 \pm 1421.76$ \\
Avg. Entities/S        & $1.24 \pm 1.82$ & $18.87 \pm 37.16$ \\
Avg. Concepts/S        & $1.24 \pm 1.82$ & $18.87 \pm 37.16$ \\
Total Entities         & \multicolumn{2}{c}{261,457} \\
Entity Types           & \multicolumn{2}{c}{5} \\
Unique Concepts        & \multicolumn{2}{c}{3,953} \\
\bottomrule
\end{tabular}
}
\end{table}

\noindent \textbf{Automated Annotation Engine.} We build a rule-based parser that reconstructs semantic links between human-readable disclosures and Inline XBRL (iXBRL) metadata. For each instance $i$, the parser extracts iXBRL fact elements (e.g., \texttt{ix:nonfraction}) and retrieves the value $v$, concept $c$, and data type $\tau$, implemented as \textsc{ParseIXBRL} in Algorithm~\ref{alg:step1}. To avoid duplicates from nested tags, we restrict extraction to atomic leaf elements $\epsilon$.

\begin{algorithm}[t]
\scriptsize
\caption{Automated Instance Annotation}
\label{alg:step1}
\KwIn{Raw iXBRL reports $\mathcal{D}$; taxonomy $\mathcal{T}$; valid types $\mathcal{L}$}
\KwOut{Positive instances $\mathcal{P}$; Negative instances $\mathcal{N}$}

Initialize $\mathcal{P}\!\gets\!\emptyset$, $\mathcal{N}\!\gets\!\emptyset$, seen text set $\mathcal{S}\!\gets\!\emptyset$\;

\ForEach{instance $i \in \mathcal{D}$ (sentence/table)}{
    $text \gets \textsc{ExtractText}(i)$\;
    \If{$|text|\le 20$ \textbf{or} $text \in \mathcal{S}$}{\textbf{continue}}
    
    $E \gets \{(v,\tau,c)\mid (v,c,\tau)\!\leftarrow\!\textsc{ParseIXBRL}(i),\ c\!\in\!\mathcal{T},\ \tau\!\in\!\mathcal{L},\ v\neq\emptyset\}$\;
    $E \gets \textsc{DedupBy}(E,\ (v,c))$\;
    
    \eIf{$E\neq\emptyset$}{
        $\mathcal{P}\gets \mathcal{P}\cup\{(text,E)\}$\;
    }{
        $\mathcal{N}\gets \mathcal{N}\cup\{(text,\emptyset)\}$\;
    }
    $\mathcal{S}\gets \mathcal{S}\cup\{text\}$\;
}
\Return $\mathcal{P}, \mathcal{N}$\;
\end{algorithm}

\begin{figure}[h]
  \centering
  \setlength{\abovecaptionskip}{2pt}
  \setlength{\belowcaptionskip}{0pt}
  \includegraphics[width=\linewidth]{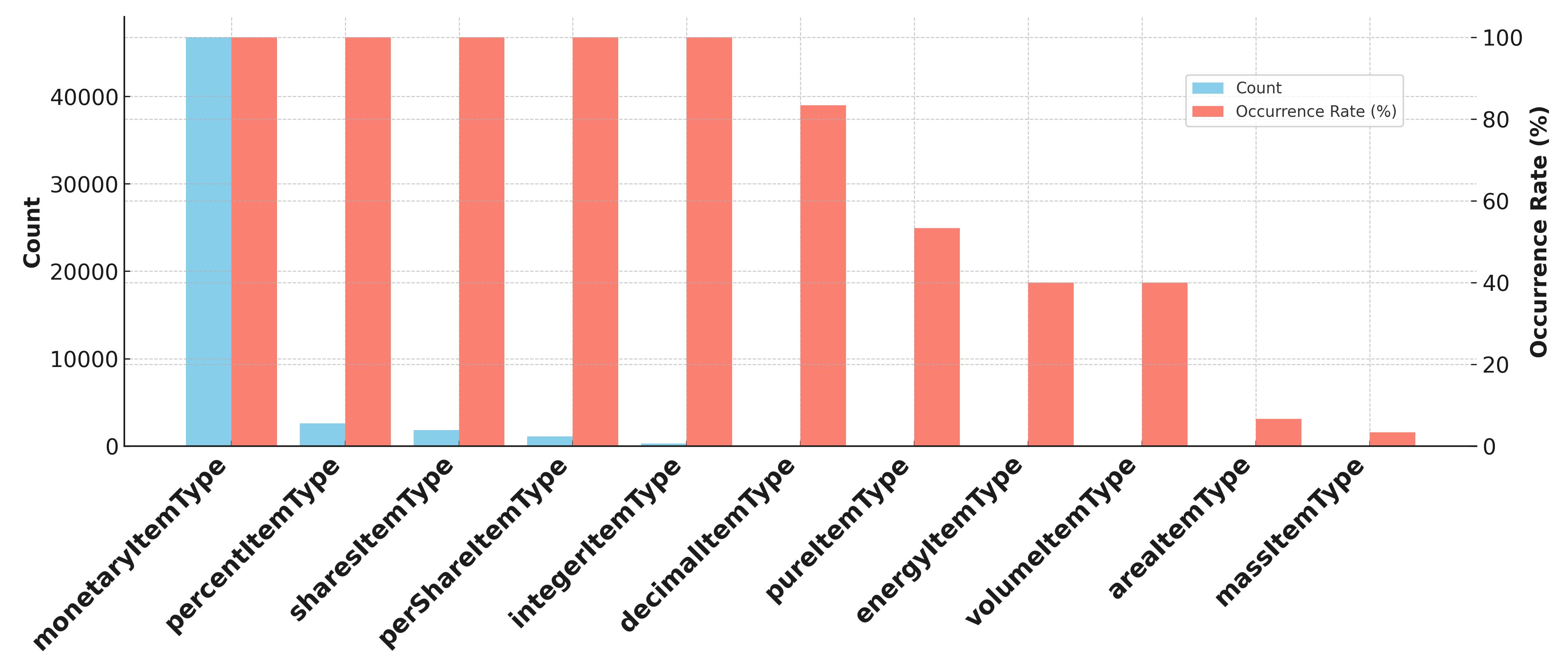}
  \caption{Statistics of numerical entity types.}
  \label{fig:statistic_entitytype}
\end{figure}

\noindent \textbf{Valid Entity Types.} To define the label space, we profile all numeric XBRL item types. As shown in Figure~\ref{fig:statistic_entitytype}, the distribution is long-tailed: although eleven types appear, five dominate. We therefore restrict $\mathcal{L}$ to \textit{monetaryItemType}, \textit{percentItemType}, \textit{sharesItemType}, \textit{perShareItemType}, and \textit{integerItemType}. An instance is labeled POSITIVE ($\mathcal{P}$) only if it contains at least one concept $c \in \mathcal{T}$ from $\mathcal{L}$; otherwise it is labeled NEGATIVE ($\mathcal{N}$).

\noindent \textbf{Structural Refinement and Filtering.} For reliable evaluation, we keep only text segments longer than \textbf{20 characters} and deduplicate content globally using $text$. For tables, we retain structural tags (\texttt{<table>}, \texttt{<tr>}, \texttt{<td>}, \texttt{<th>}) but discard tables without numerical facts. This filtering preserves semantic accuracy while maintaining realistic layout structure.

\subsection{Data Quality Control}

Validating all 261,457 entities is infeasible. We therefore adopt a strategic sampling scheme based on a greedy minimal-cover method (algorithm~\ref{alg:step2}) to construct a compact audit set $\mathcal{P}^*$. The procedure balances two objectives: \textit{Company Balance}, by capping the number of sampled instances per company, and \textit{Taxonomy Completeness}, by ensuring coverage of all observed US-GAAP concepts. Specifically, we first seed $\mathcal{P}^*$ with up to $K$ instances per company, then iteratively add instances that cover the largest number of remaining uncovered concepts. This process yields a diversity-maximized subset consisting of 928 sentence instances and 1,021 table instances, which is reviewed by two junior analysts and one senior adjudicator.

\begin{algorithm}[t]
\scriptsize
\caption{Strategic Audit Set Sampling}
\label{alg:step2}
\KwIn{Positive instances $\mathcal{P}$; observed concepts $\mathcal{C}_{all}$; cap $K=10$}
\KwOut{Strategic subset $\mathcal{P}^*$}

$\mathcal{P}^*\!\gets\!\emptyset$;\ $\mathcal{C}_{rem}\!\gets\!\mathcal{C}_{all}$\;

\ForEach{company $g$}{
    Add up to $K$ instances from $g$ to $\mathcal{P}^*$\;
    $\mathcal{C}_{rem}\gets \mathcal{C}_{rem}\setminus \textsc{Concepts}(\mathcal{P}^*)$\;
}

\While{$\mathcal{C}_{rem}\neq\emptyset$ \textbf{and} $\mathcal{P}\setminus\mathcal{P}^*\neq\emptyset$}{
    $i^*\gets \arg\max_{i\in \mathcal{P}\setminus\mathcal{P}^*} |\textsc{Concepts}(i)\cap \mathcal{C}_{rem}|$\;
    \If{ties}{select $i^*$ with the shortest text length\;}
    $\mathcal{P}^*\gets \mathcal{P}^*\cup\{i^*\}$;\ $\mathcal{C}_{rem}\gets \mathcal{C}_{rem}\setminus \textsc{Concepts}(i^*)$\;
}
\Return $\mathcal{P}^*$\;
\end{algorithm}

\noindent \textbf{Review Protocol.} The audit followed two stages. A pilot phase aligned interpretations of the \textit{Verification Guideline}\footnote{\url{https://docs.google.com/document/d/1TgTxz-fozBImeRs8Y9kLViHnr8Bjs1qZbsB8u5NjfmA/edit?usp=sharing}}, after which analysts independently verified whether extracted triplets $(v,\tau,c)$ were fully correct. The senior expert adjudicated disagreements to create the final gold set.

\noindent \textbf{Agreement Results.} On the independent round, annotators achieved \textbf{96\% raw agreement} and \textbf{Cohen’s $\kappa=0.81$}, indicating substantial agreement~\cite{landis1977measurement}. This high-verified accuracy on the most semantically diverse subset provides strong evidence for the reliability of our automated annotation pipeline.

\subsection{Task-Specific Dataset Construction}
Following validation, we integrated the corrected $\mathcal{P}^*$ subset back into the full corpus to form a high-fidelity hybrid dataset. This refined pool ensures that every US-GAAP concept in the taxonomy has at least one expert-verified representation. From this foundation, we derive two subtask-specific datasets: \textbf{FinNI-eval}, targeting numerical entity identification, and \textbf{FinCL-eval}, targeting concept linking. Both are the first datasets in the financial domain to emphasize structured information extraction and concept-level alignment. Their detailed statistics are reported in Table~\ref{tab:eval_stats}.

\begin{table}[h!]
\setlength{\abovecaptionskip}{3pt}  
\scriptsize
\centering
\caption{Statistics of evaluation datasets for FinNI-eval and FinCL-eval. 
Tokens are calculated using the ``cl100k\_base'' tokenizer ($\pm$ standard deviation).}
\label{tab:eval_stats}
\resizebox{0.8\linewidth}{!}{
\begin{tabular}{lcc}
\toprule
\textbf{Item} & \textbf{FinNI-eval} & \textbf{FinCL-eval} \\
\midrule
\#Instance         & 28,787  & 261,457 \\
Avg. Input Tokens  & $1101.50 \pm 992.00$ & $60.43 \pm 48.25$ \\
Max Input Tokens   & 18,831 & 750 \\
Avg. Output Tokens & $175.24 \pm 375.90$ & $12.82 \pm 5.09$ \\
Max Output Tokens  & 7,813  & 41 \\
\bottomrule
\end{tabular}
}
\end{table}


\subsubsection{FinNI-eval Dataset Construction}

Following the annotation procedure in Section~\ref{filtered_data}, we construct the FinNI-eval dataset with 28,787 instances derived from annotated sentences and tables. Each instance consists of an input block and an answer block. The input block includes a task instruction and the corresponding content, while the answer block is a JSON list of identified numerical values with their types (or an empty list).

\begin{tcolorbox}[colback=lightgray!10, colframe=black, title=FinNI-eval Dataset, title style={fontupper=\scriptsize}]
\scriptsize
\begin{verbatim}
Instruction: <FinNI Task Instruction>
Input: <Sentences or Table>
Answer: {"results": [{"Fact": <numeric entity>, 
        "Type": <entity type>}]} or {"results": []}
\end{verbatim}
\end{tcolorbox}

\subsubsection{FinCL-eval Dataset Construction}

Following the FinCL formulation in Section~\ref{fincl}, we construct the FinCL-eval dataset with 261,457 query–answer pairs for numerical entity normalization, along with a US-GAAP taxonomy containing 17,688 concepts. Each query includes a numerical entity, its type, and surrounding context, while the answer is the corresponding US-GAAP concept.

\begin{tcolorbox}[colback=lightgray!10, colframe=black, title=FinCL-eval Dataset,title style={fontupper=\scriptsize}]
\scriptsize
\begin{verbatim}
Query: <entity> + <entity type> + <context>
Answer: <US-GAAP Concept>
\end{verbatim}
\end{tcolorbox}

\subsection{Evaluation}
\label{eval_sec}
\begin{figure}[h]
  \centering
  \setlength{\abovecaptionskip}{2pt}
  \setlength{\belowcaptionskip}{0pt}
  \includegraphics[width=1.0\linewidth]{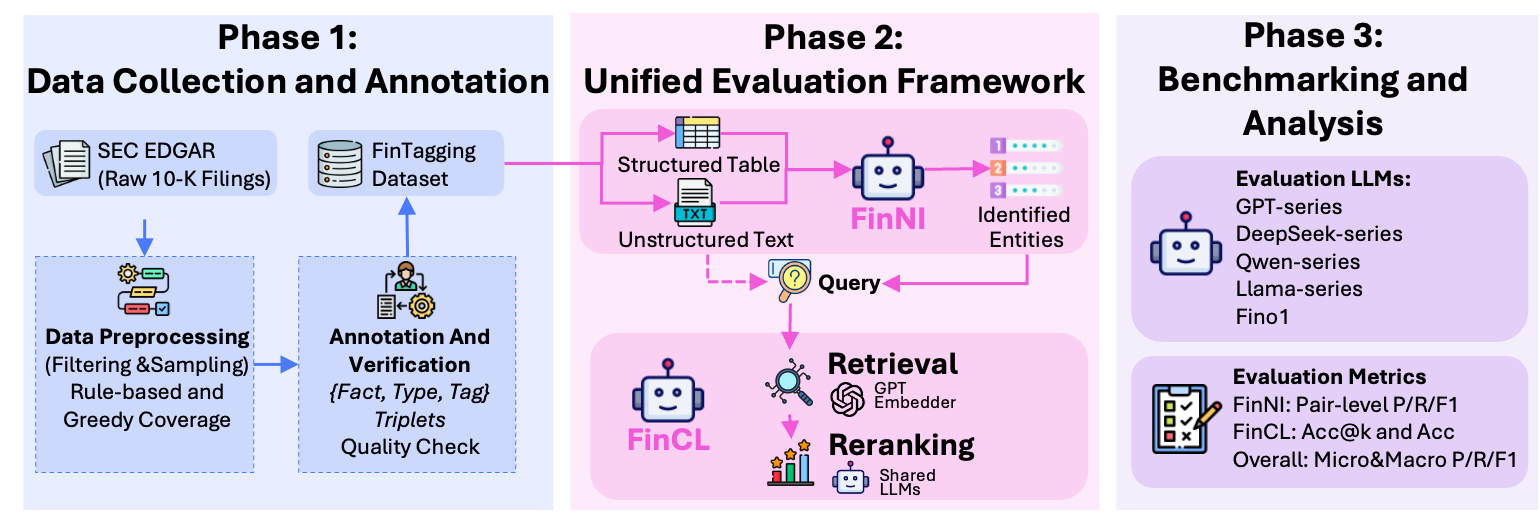}
  \caption{A unified evaluation framework on \textsc{\fintagging} benchmark. Note: The \textbf{Input content} is drawn from the \textsc{\fintagging} dataset, and the "Shared LLMs" are used for identification in the FinNI task and reranking in the FinCL task.}
  \label{fig:framework}
\end{figure}

We propose a unified evaluation framework for \textsc{\fintagging}. As shown in Figure~\ref{fig:framework}, each input is decomposed into two subtasks, which together produce structured triplets. This design jointly evaluates zero-shot numerical extraction and concept alignment. For comparison with token-classification baselines, we report macro and micro Precision, Recall, and F1~\cite{sokolova2009systematic}. FinNI is evaluated using pair-level metrics, while FinCL is evaluated with accuracy.

\textbf{\textit{FinNI Evaluation.}} FinNI uses pair-level metrics to assess whether models correctly extract numerical values from text-table inputs.

\textbf{\textit{FinCL Evaluation.}} FinCL is formulated as a retrieval–reranking task to avoid impractical extreme classification. We embed each taxonomy concept using text-embedding-3-small\footnote{\label{es}\url{https://platform.openai.com/docs/models/text-embedding-3-small}} and retrieve the top-$h$ candidates from the US-GAAP taxonomy by semantic similarity. The LLM then reranks candidates using contextual reasoning to select the final tag.

\subsubsection{Evaluation Models}

Our goal is to assess the foundational capabilities of SOTA LLMs on \textsc{\fintagging} and understand their strengths and limitations in XBRL tagging. We evaluate models across three categories using our unified framework.
(1) Closed-source general LLM: GPT-4o~\cite{hurst2024gpt}.
(2) Open-source general LLMs: DeepSeek-V3~\cite{liu2024deepseek}, DeepSeek-R1-Distill-Qwen-32B~\cite{deepseekai2025deepseekr1incentivizingreasoningcapability}, Qwen3-32B / 14B / 1.7B / 0.6B~\cite{qwen3technicalreport}, Llama-4-Scout-17B-16E-Instruct~\cite{meta2025llama}, Llama-3.3-70B-Instruct, Llama-3.1-8B-Instruct, Llama-3.2-3B-Instruct~\cite{grattafiori2024llama}, and gemma-2-27b-it~\cite{gemma_2024}.
(3) Domain-specific financial LLM: Fino1-8B~\cite{qian2025fino1}.
We also compare against strong PLM baselines: BERT-large~\cite{DBLP:journals/corr/abs-1810-04805}, FinBERT~\cite{araci2019finbert}, and SECBERT~\cite{loukas2022finer}.
Models are evaluated with the LM Evaluation Harness~\cite{eval-harness} on a 4$\times$H100 cluster and via APIs, using input limits of 24,576 (FinNI) and 8,192 (FinCL) tokens. Total cost is approximately 500 GPU-hours and \$1,500 in API usage.


\section{Experiment and Result}
\label{sec:exp}

\subsection{Overall Results}
Table~\ref{tab:overall_res} presents the overall performance on the \textsc{\fintagging} benchmark. It clearly demonstrates that under our framework, LLMs can effectively handle both frequent and rare financial tags, indicating their ability to mitigate long-tail label challenges and underscoring the advantage of our information extraction and alignment formulation over traditional token-level classification approaches. 

\begin{table}[!h]
\setlength{\abovecaptionskip}{3pt}
  \caption{Overall Performance. Bolded values indicate the best performance, underlined values represent the second-best, and italicized values denote the third-best performance.}
  \label{tab:overall_res}
  \centering
  \resizebox{.95\linewidth}{!}{\begin{tabular}{lcccccc}
    \toprule
    \multirow{2}{*}{Models} & \multicolumn{3}{c}{Macro} & \multicolumn{3}{c}{Micro}  \\ \cline{2-4} \cline{5-7}
     & P & R & F1 & P & R & F1 \\
    \midrule
    GPT-4o & \underline{0.0865} & \underline{0.0626} & \underline{0.0569} & 0.1052 & 0.0731 & 0.0863 \\
    \midrule
    DeepSeek-V3 & \textbf{0.0949} & \textbf{0.0778} & \textbf{0.0659} & 0.1074 & \textit{0.1264} & \textit{0.1162} \\
    Llama-4-Scout-17B-16E-Instruct & \textit{0.0683} & \textit{0.0414} & \textit{0.0400} & 0.1045 & 0.0526 & 0.0699 \\
    Llama-3.3-70B-Instruct & 0.0544 & 0.0279 & 0.0288 & 0.0665 & 0.0382 & 0.0485 \\
    DeepSeek-R1-Distill-Qwen-32B & 0.0532 & 0.0283 & 0.0285 & 0.0814 & 0.0214 & 0.0339 \\
    Qwen3-32B & 0.0639 & 0.0314 & 0.0324 & \textit{0.1127} & 0.0230 & 0.0382 \\
    gemma-2-27b-it & 0.0471 & 0.0291 & 0.0276 & 0.0533 & 0.0390 & 0.0451 \\
    Qwen3-14B & 0.0591 & 0.0274 & 0.0288 & 0.1069 & 0.0182 & 0.0311 \\
    Llama-3.1-8B-Instruct & 0.0345 & 0.0178 & 0.0169 & 0.0575 & 0.0166 & 0.0258 \\
    Llama-3.2-3B-Instruct & 0.0194 & 0.0118 & 0.0100 & 0.0178 & 0.0084 & 0.0114 \\
    Qwen3-1.7B & 0.0207 & 0.0067 & 0.0080 & 0.1063 & 0.0031 & 0.0060 \\
    Qwen3-0.6B & 0.0026 & 0.0008 & 0.0010 & 0.0562 & 0.0002 & 0.0005 \\
    \midrule
    Fino1-8B & 0.0344 & 0.0143 & 0.0151 & 0.0419 & 0.0128 & 0.0197 \\
    \midrule
    BERT-large & 0.0252 & 0.0266 & 0.0205 & \underline{0.1518} & \underline{0.1283} & \underline{0.1391} \\
    FinBERT & 0.0046 & 0.0064 & 0.0042 & 0.0872 & 0.0526 & 0.0656 \\
    SECBERT & 0.0231 & 0.0295 & 0.0203 & \textbf{0.1870} & \textbf{0.1697} & \textbf{0.1779} \\
    \bottomrule
  \end{tabular}}
\end{table}

From a \textbf{macro perspective}, which emphasizes balanced performance across frequent and rare tags, large general-purpose LLMs clearly dominate. DeepSeek-V3, GPT-4o, and Llama-4-Scout achieve the strongest macro-F1 scores, surpassing all fine-tuned PLMs and indicating better generalization to long-tail concepts. Qwen3-32B also performs competitively, suggesting that model architecture and pretraining can compensate for a smaller scale.
From a \textbf{micro perspective}, which reflects performance on frequent labels, the trend is similar: top LLMs remain competitive even without fine-tuning, while fine-tuned PLMs such as SECBERT and BERT-large excel mainly because they benefit from label frequency rather than broader coverage.

\subsection{Subtask Results}

Table~\ref{tab:finni_fincl_res} summarizes performance on the FinNI and FinCL subtasks. Overall, large LLMs outperform smaller models across both tasks in a zero-shot setting. On FinNI, DeepSeek-V3 achieves the best F1 score, driven by substantially higher recall, while GPT-4o performs competitively. In contrast, several models (e.g., Qwen3 variants) exhibit high precision but extremely low recall, indicating that they extract only a small fraction of target facts, which leads to poor F1 performance. Mid-sized models such as gemma-2-27B show more balanced behavior. In contrast, smaller models and the financial LLM Fino1-8B lag, suggesting that domain pretraining alone does not yield robust gains for numerical extraction.

\begin{table}[!h]
\setlength{\abovecaptionskip}{3pt}
\caption{Performance comparison of different models on the FinNI and FinCL subtasks.
Bolded values indicate the best performance, underlined values represent the second-best, and italicized values denote the third-best performance.}
\label{tab:finni_fincl_res}
\centering
\resizebox{.85\linewidth}{!}{%
\begin{tabular}{lcccc}
\toprule
\multirow{2}{*}{Model} & \multicolumn{3}{c}{FinNI} & FinCL \\
\cline{2-4} \cline{5-5}
 & Precision & Recall & F1 & Accuracy \\
\midrule
GPT-4o & 0.5893 & \underline{0.4977} & \underline{0.5397} & \underline{0.1829} \\
\midrule
DeepSeek-V3 & 0.5886 & \textbf{0.8430} & \textbf{0.6932} & \textbf{0.1889} \\
Llama-4-Scout-17B-16E-Instruct & 0.4668 & 0.3164 & 0.3771 & \textit{0.1649} \\
Llama-3.3-70B-Instruct & 0.4826 & 0.3301 & 0.3920 & 0.1318 \\
DeepSeek-R1-Distill-Qwen-32B & 0.5676 & 0.1942 & 0.2894 & 0.1141 \\
Qwen3-32B & \underline{0.6991} & 0.1804 & 0.2868 & 0.1277 \\
gemma-2-27b-it & 0.5060 & \textit{0.4526} & \textit{0.4778} & 0.1099 \\
Qwen3-14B & \textit{0.6912} & 0.1487 & 0.2448 & 0.1144 \\
Llama-3.1-8B-Instruct & 0.3874 & 0.1761 & 0.2421 & 0.0913 \\
Llama-3.2-3B-Instruct & 0.1856 & 0.1203 & 0.1460 & 0.0415 \\
Qwen3-1.7B & \textbf{0.7362} & 0.0281 & 0.0541 & 0.0735 \\
Qwen3-0.6B & 0.2803 & 0.0019 & 0.0038 & 0.0414 \\
\midrule
Fino1-8B & 0.3431 & 0.1293 & 0.1878 & 0.0807 \\
\bottomrule
\end{tabular}%
}
\end{table}

FinCL remains significantly more challenging than FinNI. All FinCL results are reported under a retrieval–reranking setup, where models select the final tag from the top-200 retrieved US-GAAP candidates. Although retrieval results are not separately tabulated due to space constraints, structure-aware retrieval achieves over 0.65 Acc@200 for sentence-based entities but remains below 0.30 Acc@200 for table-based entities, resulting in an overall Acc@200 of approximately 0.33. This indicates that retrieval coverage is already limited, especially for tabular contexts, which imposes a ceiling on downstream reranking. Even under this constrained candidate space, the best-performing models achieve accuracy below 0.19, with DeepSeek-V3 and GPT-4o forming a narrow top tier. These results highlight a clear extraction–alignment gap: while modern LLMs can recover numerical facts with reasonable accuracy, fine-grained US-GAAP disambiguation remains intrinsically difficult, particularly for table-origin entities.

\subsection{Ablation analysis}
We further compare our benchmark against extreme multi-class classification settings. We select the best-performing model from each category. The evaluation uses triplet-level ($\{\texttt{Tag}, \texttt{Fact}, \texttt{Type}\}$) Precision, Recall, and F1.

\begin{table}[!h]
\setlength{\abovecaptionskip}{3pt}  
\scriptsize
\caption{Performance comparison between with/without our evaluation framework on the \textsc{\fintagging} benchmark dataset.}
\label{tab:comparison_framework}
\centering
\resizebox{.85\linewidth}{!}{%
\begin{tabular}{lllll}
    \toprule
    Evaluation Mode & Model & Precision & Recall & F1 \\
    \midrule
    \multirow{3}{*}{FINTAGGING} 
      & GPT-4o & 0.1166 & 0.0915 & 0.1026 \\
      & DeepSeek-V3 & 0.1187 & 0.1581 & 0.1356 \\
      & Fino1-8B & 0.0336 & 0.0140 & 0.0197 \\
    \midrule
    \multirow{3}{*}{Extreme Classification} 
      & GPT-4o & 0 & 0 & 0 \\
      & DeepSeek-V3 & 0 & 0 & 0 \\
      & Fino1-8B & 0 & 0 & 0 \\
    \bottomrule
\end{tabular}%
}
\end{table}

Table~\ref{tab:comparison_framework} shows that modeling XBRL tagging as single-step extreme classification fails in practice: models must select from thousands of taxonomy concepts without grounding in identified numerical entities, leading all models to collapse to zero. In contrast, our two-stage framework reflects real tagging workflows: FinNI first identifies and types numerical facts, and FinCL then selects from a retrieved candidate set using context. This formulation yields non-trivial performance and exposes meaningful differences among models that are entirely obscured under extreme classification.

\subsection{Error propagation in \textsc{\fintagging} pipeline}

Table~\ref{tab:pipeline_error} compares accuracy under three settings: the full pipeline, FinCL with gold entities, and rerank-only with gold candidates. Results show clear error cascading: end-to-end performance is lowest due to accumulated extraction and retrieval errors. Providing gold entities substantially improves accuracy, identifying entity extraction as a key bottleneck. Rerank-only settings further increase performance, yet accuracy remains low, indicating that fine-grained US-GAAP disambiguation is inherently challenging even with perfect retrieval.


\begin{table}[t]
\setlength{\abovecaptionskip}{3pt}  
\scriptsize
\centering
\caption{Accuracy comparison across the full pipeline, FinCL subtask, and rerank-only settings.}
\label{tab:pipeline_error}
\resizebox{.85\linewidth}{!}{%
\begin{tabular}{lccc}
\toprule
Model & Pipeline & FinCL & Rerank-only \\
\midrule
GPT-4o & 0.1052 & 0.1829 & 0.2023 \\
DeepSeek-V3  & 0.1074 & 0.1889 & 0.2144 \\
Llama-4-Scout-17B-16E-Instruct & 0.1045 & 0.1649 & 0.1792 \\
Llama-3.3-70B-Instruct  & 0.0665 & 0.1318 & 0.1482\\
DeepSeek-R1-Distill-Qwen-32B & 0.0814 & 0.1141 & 0.1258\\
Qwen3-32B & 0.1127 & 0.1277 & 0.1402\\
gemma-2-27b-it & 0.0533 & 0.1099 & 0.1210\\
Qwen3-14B & 0.1069 & 0.1144 & 0.1321\\
Llama-3.1-8B-Instruct & 0.0575 & 0.0913 & 0.1076\\
Llama-3.2-3B-Instruct & 0.0178 & 0.0415 & 0.0566\\
Qwen3-1.7B & 0.0598 & 0.0735 & 0.0880\\
Qwen3-0.6B & 0.0362 & 0.0414 & 0.0533\\
Fino1-8B   & 0.0419 & 0.0807 &  0.0908\\
\bottomrule
\end{tabular}%
}
\end{table}

\section{Conclusion}
This paper introduces \textsc{\fintagging}, a benchmark for evaluating LLMs on XBRL tagging of real financial reports. The task is structured into two subtasks, Financial Numerical Identification and Financial Concept Linking, to separately assess extraction and concept alignment. Our results show that LLMs perform reasonably well in zero-shot settings and generalize to long-tail entities, but still struggle to align facts with precise US-GAAP concepts. These findings point to gaps in structure-aware and semantic reasoning. \textsc{\fintagging} provides a foundation for future work on reliable XBRL tagging and regulatory reporting. 
\bibliographystyle{ACM-Reference-Format}
\bibliography{custom}

\clearpage
\appendix

\section{Significance analysis}
\label{significance}

\begin{figure}[!h]
  \centering
  \begin{subfigure}[t]{0.48\textwidth}
    \centering
    \includegraphics[width=\linewidth]{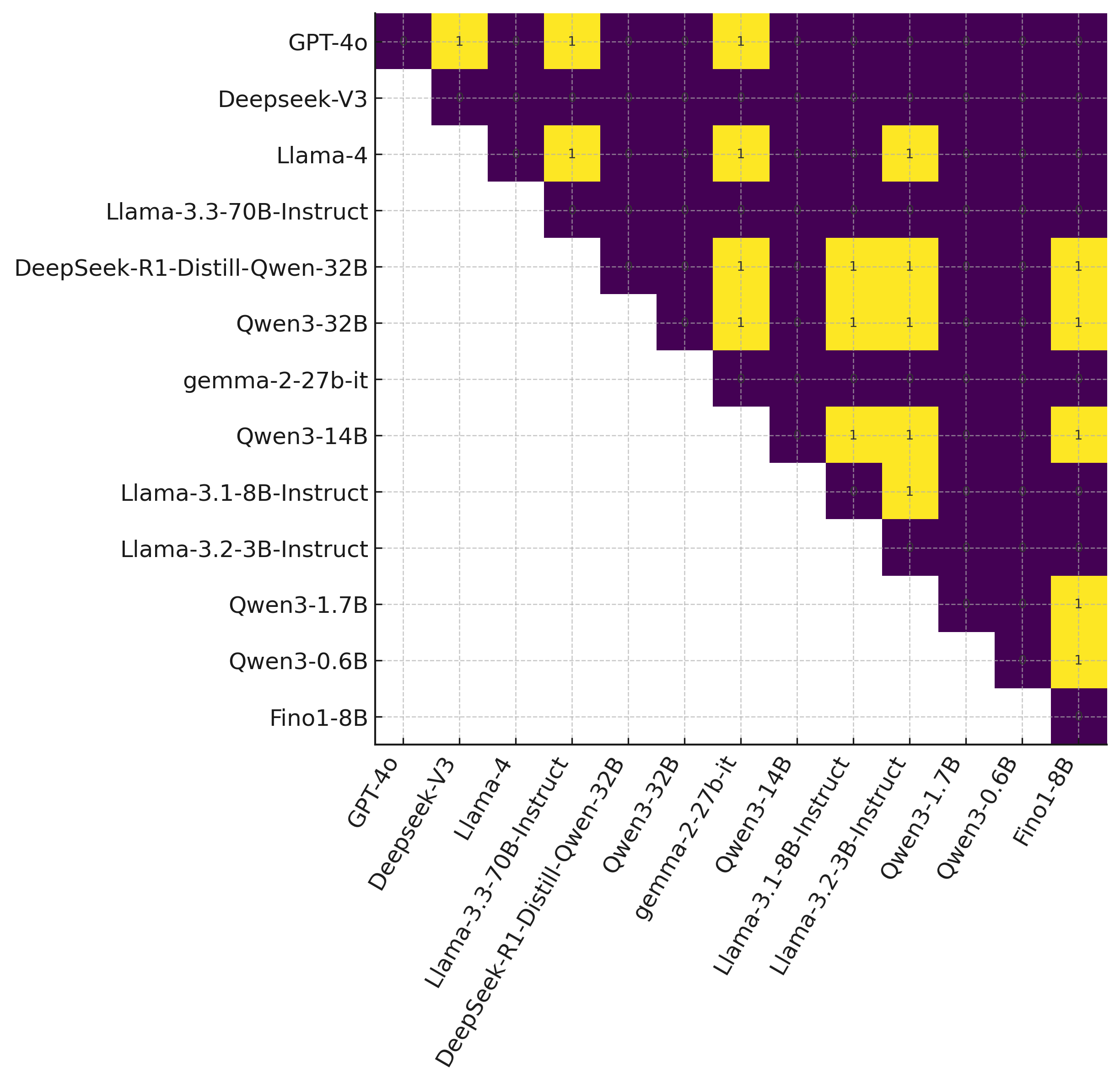}
    \caption{Model-vs-Model significance matrix.}
    \label{fig:sig_matrix}
  \end{subfigure}\hfill
  \begin{subfigure}[t]{0.48\textwidth}
    \centering
    \includegraphics[width=\linewidth]{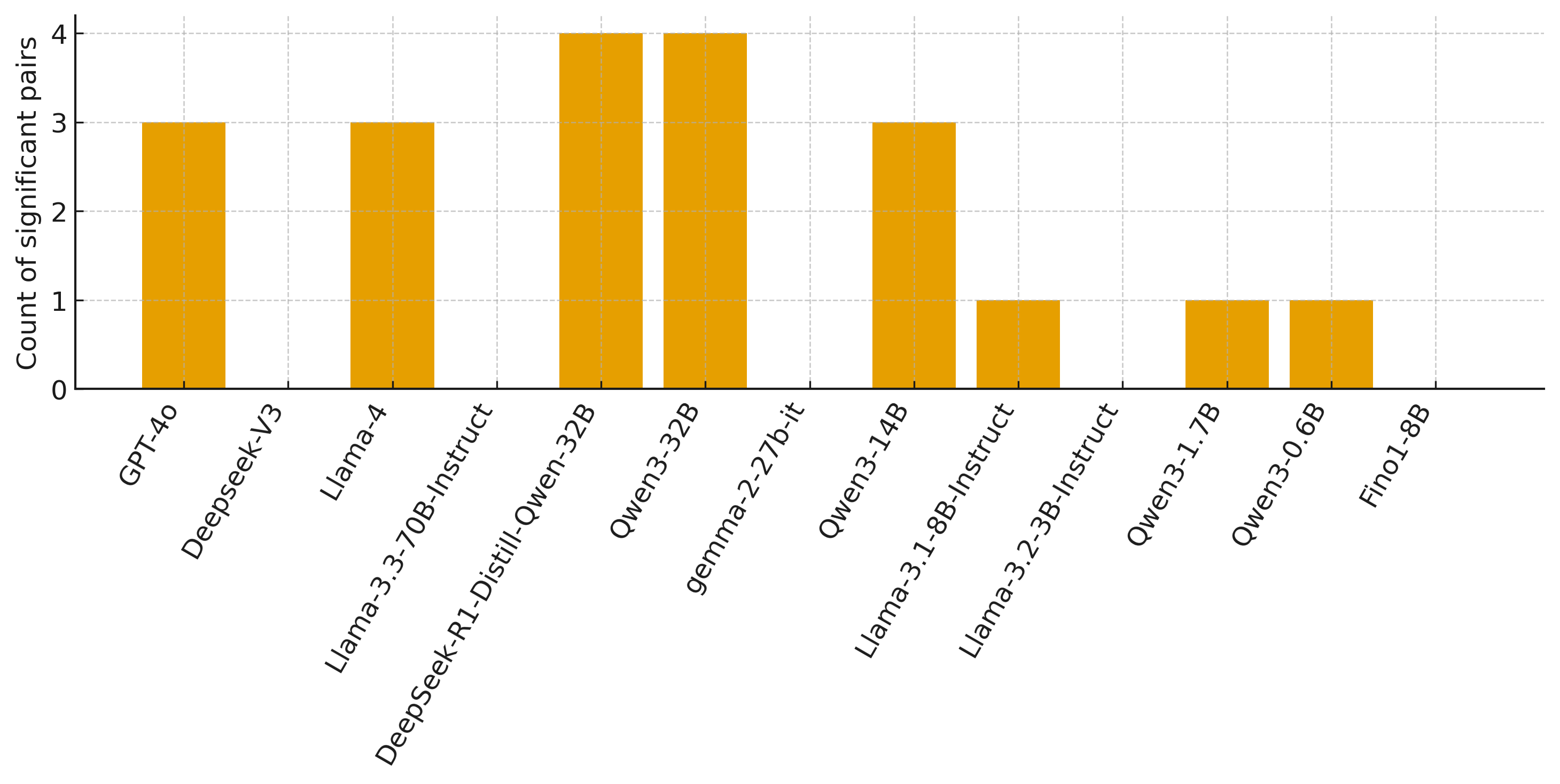}
    \caption{Per-model count of significant pairwise differences.}
    \label{fig:sig_counts}
  \end{subfigure}
  \caption{Visualization of significance analysis across models. (a) shows the pairwise-bootstrap significance matrix, while (b) summarizes how many significant differences each model has relative to others.}
  \label{fig:significance_overview}
\end{figure}

To further examine the comparative performance across models, we conducted a pairwise significance analysis, with the results summarized in Figure~\ref {fig:significance_overview}. The upper-triangular matrix in SubFigure~\ref{fig:sig_matrix} presents the pairwise significance outcomes, where a value of 1 indicates that the row model’s performance is significantly different from the column model. In contrast, 0 indicates that the difference is not statistically significant. The diagonal elements are set to zero, since self-comparisons are not meaningful.

As shown in SubFigure~\ref{fig:sig_counts}, the number of significant pairwise differences varies across models, reflecting heterogeneous performance behaviors on the evaluated benchmark. The models DeepSeek-R1-Distill-Qwen-32B and Qwen3-32B exhibit the largest numbers of significant differences, followed by GPT-4o, Llama-4, and Qwen3-14B, suggesting that their performance diverges more noticeably from other models. These results may indicate that models within the Qwen family display stronger variability across comparisons, potentially due to differences in training objectives or architectural configurations. In contrast, DeepSeek-V3, Llama-3.3-70B-Instruct, gemma-2-27b-it, and Fino1-8B demonstrate relatively fewer significant differences, implying that their performance remains more stable and comparable to other high-performing models. Overall, these findings suggest that larger or instruction-tuned models tend to produce more consistent outcomes, while models with smaller sizes or domain-oriented tuning show greater variation and clearer statistical separations under significance testing.

Overall, this analysis highlights not only which models outperform or underperform others but also which models behave similarly when subjected to statistical evaluation. The pairwise significance matrix, therefore, provides an interpretable view of model robustness and performance distinctiveness beyond average accuracy metrics.

\section{Literature Review}
\label{existing_works}

The XBRL provides a comprehensive taxonomy for financial reporting, encompassing thousands of detailed tags corresponding to concepts within financial statements. Applying NER to assign XBRL tags is an emerging yet challenging area.

\subsection{XBRL Tagging benchmark}
FiNER systematically benchmarked several neural architectures on the finer-139 dataset to address numeric-heavy XBRL tagging~\cite{loukas2022finer}. The initial experiments showed that standard BERT underperforms due to subword fragmentation, then the authors introduced pseudo-token strategies replacing numerals with [NUM] or [SHAPE] tokens to stabilize label assignment across fragmented numeric spans. These strategies, combined with domain-specific pretraining on SEC-BERT, significantly improved tagging performance, reaching 82.1 micro-F1 without the need for computationally expensive CRF layers. Their experiments demonstrated that subword-aware models with numeric-aware pseudo-tokens outperform both word-level BiLSTMs and vanilla BERT, particularly in numeric-heavy contexts, and avoid nonsensical label sequences. FNXL extended this benchmarking paradigm to a much larger label space of 2,794 US-GAAP tags, reframing the task as an extreme classification problem~\cite{sharma2023financial}. They compared the FiNER sequence labeling approach with a two-step pipeline that first identifies numeric spans and then assigns labels using AttentionXML. While FiNER achieved stronger micro-F1 (75.84), reflecting better performance on frequent tags, AttentionXML outperformed FiNER in macro-F1 (47.54), highlighting its strength in predicting infrequent, tail-end labels. FNXL further evaluated both models under a Hits@k setting, confirming that label recommendations from the AttentionXML pipeline could substantially reduce manual effort and maintain high inter-annotator agreement. Together, these benchmarks reveal the need for context-aware reasoning and label-ranking mechanisms in realistic XBRL tagging scenarios.

\subsection{XBRL Tagging Methods}
The previous studies also explored approaches to address scalability, semantic ambiguity, and reasoning gaps in XBRL tagging to improve performance. Saini et al.~\cite{saini2021galaxc} proposed GalaXC is a graph-based extreme classification framework that jointly learns over document-label graphs with per-label attention across multi-hop neighborhoods. By integrating label metadata and transitive label correlations, GalaXC outperformed leading deep classifiers by up to 18\% in micro-F1 on standard benchmarks and achieved 25\% gains in warm-start scenarios where partial labels are available. Moreover, Wang et al.~\cite{wang2023standardizing} addressed the practical challenge of custom tag standardization through a semantic similarity pipeline that leverages TF-IDF, Word2Vec, and FinBERT embeddings. Although unsupervised, the method was tested across nearly 200,000 custom tags from SEC filings between 2009 and 2022, and showed strong alignment performance, with vector-based mappings identifying viable standard tag candidates for a substantial proportion of non-compliant elements, offering a low-cost and interpretable solution for downstream financial analysis.
Shifting focus from classification to comprehension, XBRL-Agent evaluated the capabilities of large language models to reason over full XBRL reports~\cite{han2024xbrlagent}. The authors introduced two task types, domain taxonomy understanding and numeric reasoning, and found that base LLMs often hallucinated or misinterpreted financial content. To overcome these issues, XBRL-Agent incorporated retrieval-augmented generation (RAG) and symbolic calculators within an LLM-agent framework. The enhanced system achieved a 17\% accuracy gain on domain query tasks and a 42\% boost on numeric reasoning queries compared to base LLMs, validating the utility of modular tool augmentation. These improvements enabled reliable multi-step reasoning over complex disclosures such as debt instruments and derivative gains, which are difficult to capture using span-level classifiers. Collectively, these works broaden the methodological landscape of XBRL tagging from graph-based label propagation and embedding-based normalization to LLM-driven report analysis and point to a hybrid future where structured priors and reasoning tools jointly support accurate, scalable financial information extraction.
\subsection{Financial Evaluation Benchmarks}
In parallel to XBRL-specific advances, the financial NLP community has developed comprehensive benchmarks to assess broader capabilities in information extraction, numerical reasoning, and document understanding. FiNER-ORD~\cite{shah2023finer} introduced a high-quality, domain-specific NER dataset annotated over financial news, emphasizing general entity types like persons, organizations, and locations. While not numerically focused like FiNER-139, it highlights the lexical diversity of financial discourse and establishes a strong baseline for testing pretrained and zero-shot LLMs in real-world financial NER scenarios. FinQA~\cite{chen2021finqa} pushed toward explainable QA by pairing expert-written questions with annotated multi-step reasoning programs derived from earnings reports. ConvFinQA~\cite{chen2022convfinqa} extended this challenge to conversational contexts, simulating real-world question flows over sequential financial queries. TAT-QA~\cite{zhu2021tat} focused on hybrid tabular-text reasoning and required models to align cell values and document narratives, often involving aggregation, comparison, and unit-scale interpretation. Pixiu~\cite{xie2023pixiu} introduced a broader evaluation framework by releasing FinMA, a financial LLM instruction-tuned across five tasks, and assessing it on a new benchmark covering sentiment classification, QA, summarization, NER, and stock prediction. BizBench~\cite{koncel2023bizbench} framed financial QA as program synthesis over realistic, multi-modal contexts, integrating reasoning, code generation, and domain knowledge into a single evaluation pyramid. While these benchmarks highlight the growing ability of models to integrate structured and unstructured financial data, they overlook taxonomy-driven fact alignment and do not support the structured output formats required for XBRL tagging.

\section{Data and Ticker Information}
\label{data_ticker}

\subsection{The Statistics of the Tickers}
\label{tickers}

\begin{figure}
    \centering
    \includegraphics[width=0.75\linewidth]{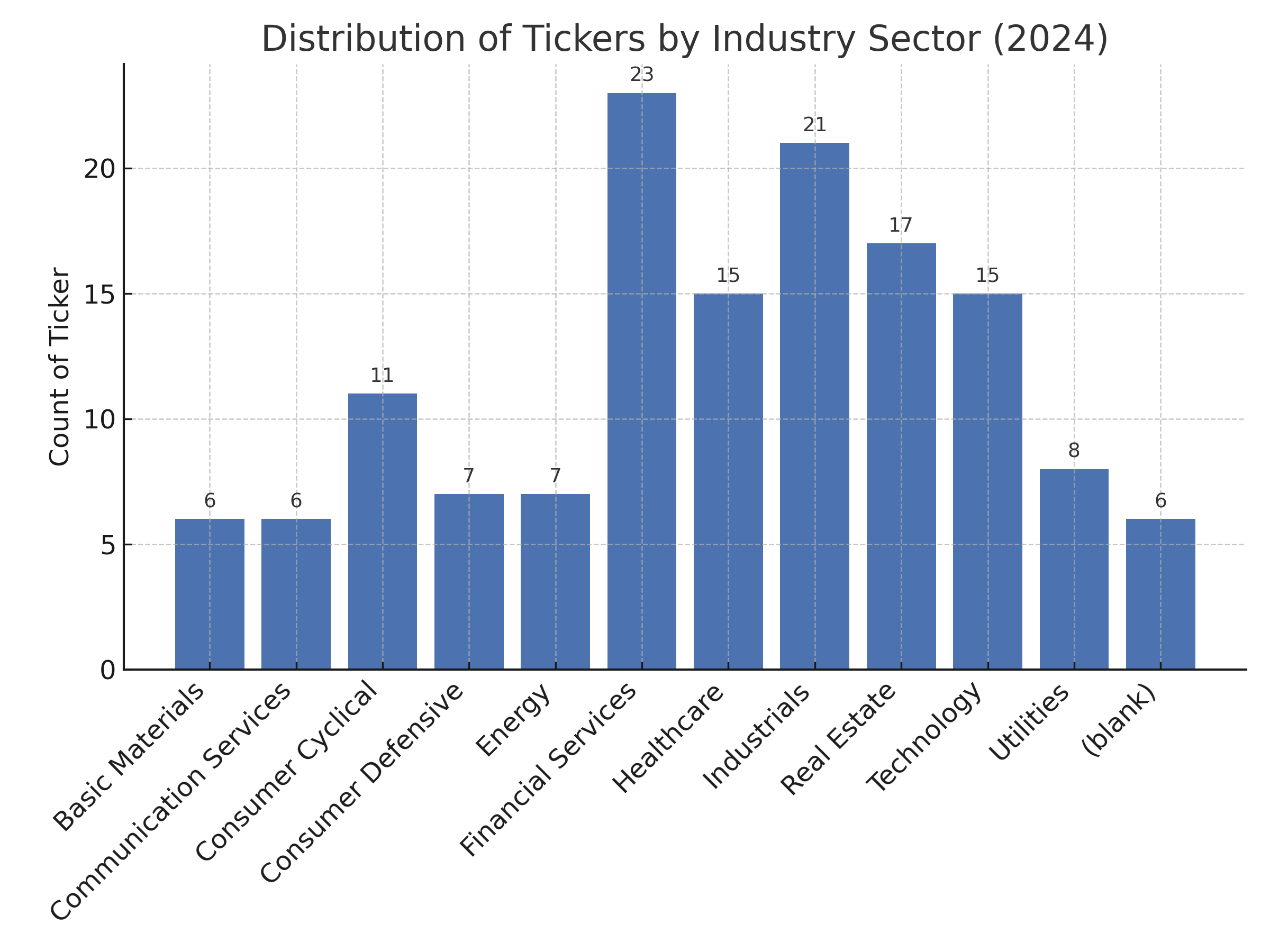}
    \caption{Distribution of Tickers by Industry Sector.}
    \label{fig:industry_sector}
\end{figure}

\begin{figure}
    \centering
    \includegraphics[width=0.6\linewidth]{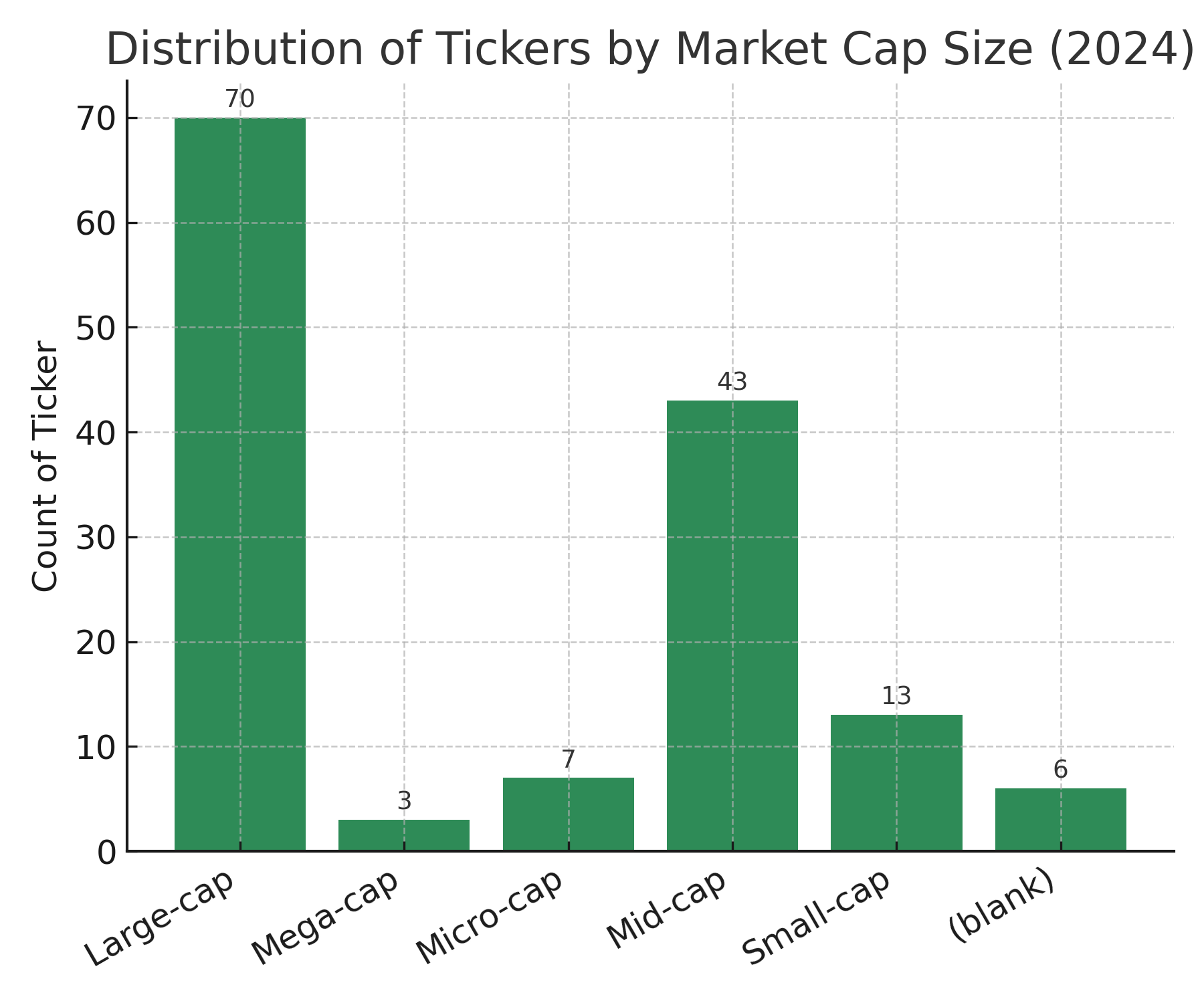}
    \caption{Distribution of Tickers by Market Cap Size.}
    \label{fig:market_cap}
\end{figure}

\begin{figure}
    \centering
    \includegraphics[width=0.6\linewidth]{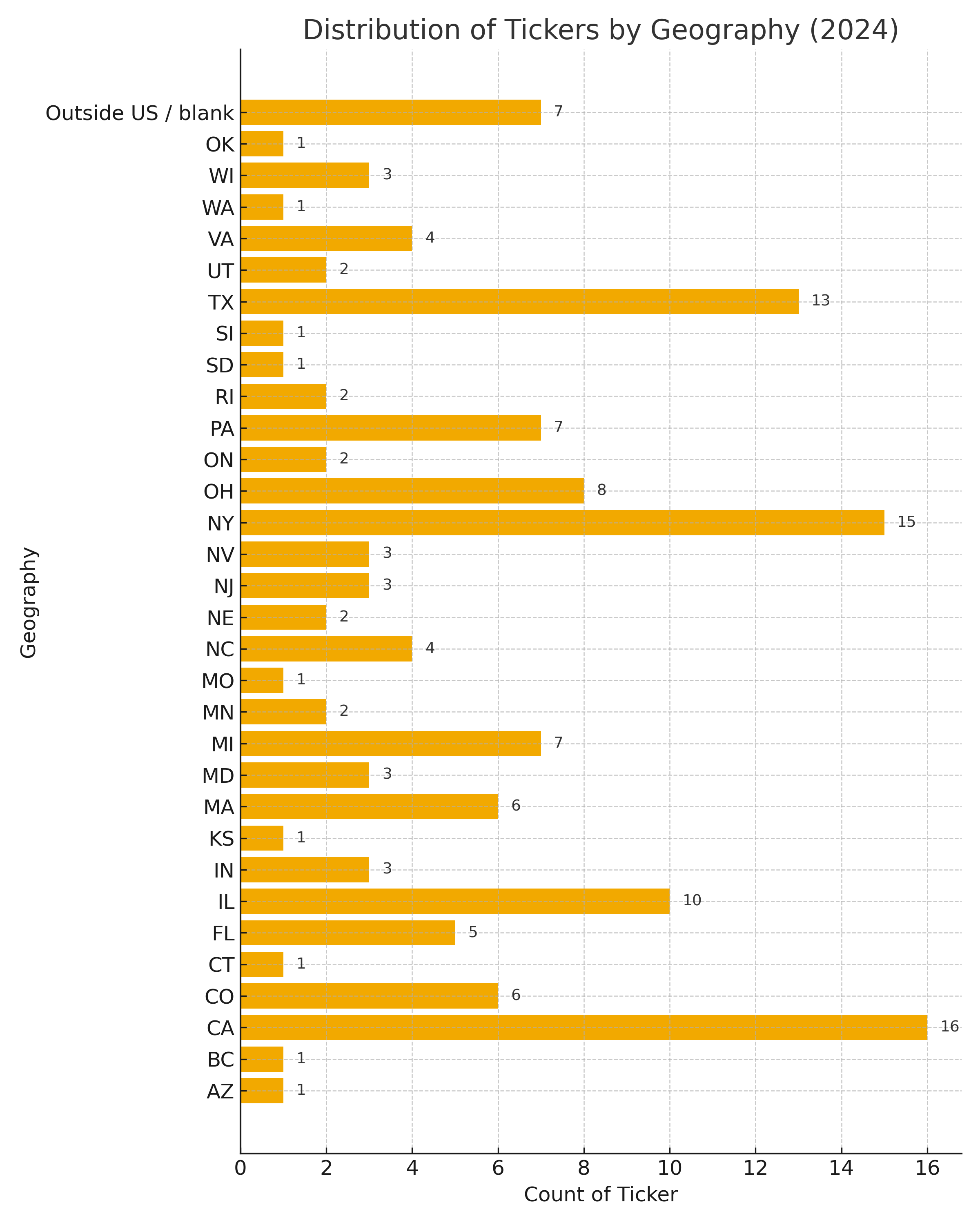}
    \caption{Distribution of Tickers by Geography.}
    \label{fig:ticker_geography}
\end{figure}

Figure~\ref{fig:industry_sector}, Figure~\ref{fig:market_cap}, and Figure~\ref{fig:ticker_geography} illustrate the distribution of 142 tickers across industry sectors, market capitalization categories, and geographic regions, highlighting the diversity of our collected financial reports. Considering that practical XBRL tagging practices often vary by industry, company size, and legal jurisdiction, we curated a diverse set of companies covering all 11 major industry sectors. The sample maintains a balanced distribution across firm sizes, following market capitalization categories: micro-cap (<\$300M), small-cap (\$300M–\$2B), mid-cap (\$2B–\$10B), large-cap (\$10B–\$200B), and mega-cap (>\$200B). In addition, we incorporated firms from over 30 states and international jurisdictions to capture regional differences in reporting and tagging conventions. This diversity ensures that the benchmark reflects realistic heterogeneity observed in financial disclosures across sectors, scales, and regulatory environments.

\subsection{Data Granularity Justification}
\label{sec:granularity}

Each 10-K report follows the SEC’s XBRL filing standard, where all narrative and tabular disclosures are linked through machine-readable tags defined in the US-GAAP taxonomy. 
While the XBRL framework provides document-level structure via instance, schema, and linkbase files, the tagging of financial values and concepts occurs at a much finer granularity—typically within individual sentences or table rows that contain numeric facts and their contextual descriptions. 

For this reason, our benchmark adopts the \textbf{sentence} and \textbf{table sequence} as the fundamental input unit rather than entire reports. 
This design choice is motivated by three considerations:

\begin{itemize}
    \item \textbf{Locality of Semantic Scope:} Financial facts and their US-GAAP tags are usually expressed within a single sentence or a bounded table region. Modeling at this level aligns the input scope with the tagging scope, reducing noise from unrelated content within long filings.

    \item \textbf{Computational Efficiency:} Full 10-K filings often exceed hundreds of pages and contain heterogeneous sections (e.g., risk factors, management discussion, and financial statements). Processing the entire report as one input sequence would exceed model context limits and obscure local numerical relations. Segment-level inputs enable fine-grained supervision and efficient evaluation.

    \item \textbf{Faithful Tag Alignment:} Because each numerical entity is explicitly linked to its taxonomy tag in the XBRL instance document, using sentence or table segments ensures a one-to-one correspondence between input content and ground-truth annotations. This granularity preserves the original tagging fidelity of the filings.
\end{itemize}

Overall, this segmentation strategy allows our benchmark to remain consistent with how XBRL tagging operates in practice (on localized factual statements), while maintaining scalability and interpretability across large collections of filings.

\subsection{Automated Data Annotation}
\label{auto_data}

Algorithm~\ref{alg:step1} converts Inline XBRL (iXBRL) filings into structured instances for our benchmark. The procedure operates over both narrative sentences and tables, and assigns each instance a positive or negative label based on whether it contains at least one validated financial entity.

We begin by iterating through all disclosure instances in the corpus and extracting their plain-text content. Very short fragments (length $\le 20$) and duplicated passages are discarded using a global text set, ensuring that the benchmark contains unique and meaningful contexts.

For each remaining instance, the algorithm traverses all atomic iXBRL elements. From each element, we retrieve the numerical value $v$, its associated concept $c$ from the US-GAAP taxonomy, and the declared data type $\tau$. We retain only elements whose concept is in $\mathcal{T}$ and whose type belongs to the predefined valid numeric set $\mathcal{L}$. Restricting extraction to leaf-level elements avoids double counting values that appear in nested XML structures. A local deduplication set further prevents repeated annotation of identical $(v,c)$ pairs within the same instance.

If at least one valid entity is identified, the instance is recorded as a positive example together with its entity list; otherwise, it is stored as a negative example. Negative instances are intentionally preserved, as they reflect realistic disclosure text that models must learn to ignore. Overall, this process yields a corpus that is faithful to regulatory metadata while remaining structurally clean and suitable for downstream evaluation.

\begin{algorithm}[t]
\footnotesize
\caption{Automated Instance Annotation}
\label{alg:step1}
\KwIn{Raw 10-K reports (iXBRL) $\mathcal{D}$; US-GAAP taxonomy $\mathcal{T}$; Set of valid data types $\mathcal{L}$}
\KwOut{Positive instances $\mathcal{P}$; Negative instances $\mathcal{N}$}

\BlankLine
Initialize: $\mathcal{P} \gets \emptyset$, $\mathcal{N} \gets \emptyset$, Global unique text set $\mathcal{S}_{text} \gets \emptyset$\;

\ForEach{instance (sentence or table) $i \in \mathcal{D}$}{
    $text \gets$ Extract plain text content from instance $i$\;
    
    \If{length of $text \le 20$ \textbf{or} $text \in \mathcal{S}_{text}$}{
        \textbf{skip} to the next instance\;
    }
    
    $E \gets \emptyset$ \tcp*[f]{List of validated entities in $i$}\;
    $S_{fact} \gets \emptyset$ \tcp*[f]{Local fact deduplication set}\;

    \ForEach{atomic iXBRL element $\epsilon$ within instance $i$}{
        Retrieve value $v$, financial concept $c$, and data type $\tau$ from $\epsilon$\;
        
        \If{$c \in \mathcal{T}$ \textbf{and} $\tau \in \mathcal{L}$ \textbf{and} $v$ is not empty}{
            \If{$(v, c) \notin S_{fact}$}{
                Add $(v, \tau, c)$ to the entity list $E$\;
                Add $(v, c)$ to the local deduplication set $S_{fact}$\;
            }
        }
    }

    \If{$E$ is not empty}{
        Add $(text, E)$ to the positive set $\mathcal{P}$\;
    }
    \Else{
        Add $(text, \emptyset)$ to the negative set $\mathcal{N}$\;
    }
    Add $text$ to the global set $\mathcal{S}_{text}$\;
}
\Return $\mathcal{P}, \mathcal{N}$\;
\end{algorithm}

\section{Annotation Data Quality Control}
\subsection{Validation Data Filtering}
\label{val_control}

To establish a reliable upper bound on annotation quality, we construct a
strategic audit subset rather than validating the full corpus. 
Algorithm~\ref{alg:step2} formalizes the sampling process as a greedy
minimal-cover problem over observed US-GAAP concepts.

The algorithm takes as input the set of positive instances $\mathcal{P}$,
together with the set of all taxonomy concepts observed in the corpus,
$\mathcal{C}_{all}$. The goal is to produce a compact subset
$\mathcal{P}^*$ that is both diverse and concept-complete.

\textbf{Seeding for company diversity.}
We first iterate over companies and select up to $K$ instances per company.
This prevents the audit set from being dominated by firms with longer filings
or more frequent disclosures. After each selection, we update the set of
uncovered concepts $\mathcal{C}_{rem}$ by removing concepts already represented
among the seeds.

\textbf{Greedy minimal cover.}
We then repeatedly add instances that contribute maximal marginal concept
coverage. For each candidate instance, we compute the number of uncovered
concepts it contains and select the instance with the largest score.
When ties occur, we prefer shorter instances to reduce annotation effort while
retaining conceptual value. After adding each instance, we update
$\mathcal{C}_{rem}$ accordingly and continue until either all concepts are
covered or no additional candidates remain.

\textbf{Resulting audit subset.}
This procedure yields a compact subset that balances two competing objectives:
coverage of the observed taxonomy and representation across companies. The
resulting $\mathcal{P}^*$ therefore serves as a stress-test sample that is both
manageable for human review and representative of the broader benchmark space.

\begin{algorithm}[t]
\small
\caption{Strategic Audit Set Sampling}
\label{alg:step2}
\KwIn{Positive instances $\mathcal{P}$ from Algorithm 1; Observed taxonomy concepts $\mathcal{C}_{all}$; Company cap $K=10$}
\KwOut{Strategic audit subset $\mathcal{P}^*$}

\BlankLine
Initialize: $\mathcal{P}^* \gets \emptyset$\;
Initialize: $\mathcal{C}_{rem} \gets \mathcal{C}_{all}$ \tcp*[f]{Set of concepts yet to be covered}\;

\noindent \textbf{Step 1: Diversity-Preserving Seeding} \\
\ForEach{unique company in $\mathcal{P}$}{
    Select up to $K$ instances belonging to this company and add to $\mathcal{P}^*$\;
    Update $\mathcal{C}_{rem}$ by removing concepts covered by these seed instances\;
}

\noindent \textbf{Step 2: Greedy Minimal Cover} \\
\While{$\mathcal{C}_{rem}$ is not empty \textbf{and} candidates remain in $\mathcal{P}$}{
    \ForEach{candidate instance $i \in \mathcal{P} \setminus \mathcal{P}^*$}{
        Calculate $Score(i) = | \text{Concepts in } i \cap \mathcal{C}_{rem} |$\;
    }
    Select instance $i^*$ with the highest $Score(i)$\;
    \If{multiple instances have the same $Score(i^*)$}{
        Break ties by selecting the instance with the shortest text length\;
    }
    Add $i^*$ to $\mathcal{P}^*$ and update $\mathcal{C}_{rem}$ by removing newly covered concepts\;
}

\Return $\mathcal{P}^*$\;
\end{algorithm}

\subsection{Annotator Demography}
\label{app_demography}
To ensure reliable verification of financial entities and taxonomy mappings, all audits were conducted by annotators with prior exposure to XBRL-based reporting and financial statement analysis. The review team consisted of two junior analysts and one senior adjudicator.

The two junior analysts are graduate-level researchers with training in accounting and financial data analytics. Both have received formal instruction on US-GAAP concepts, XBRL tagging conventions, and common disclosure practices in SEC filings. They have prior experience working with financial datasets in academic or industry projects, including tasks involving concept normalization, document review, and regulatory text interpretation. Before participating in the audit, each analyst completed a guided calibration session using our Verification Guideline to ensure consistent treatment of numerical entities, metadata attributes, and taxonomy semantics.

The senior adjudicator has over a decade of experience in financial reporting analysis and structured disclosure systems. Their professional background includes work with XBRL instances, taxonomy structure, and validation procedures used in regulatory environments. In addition to overseeing the audit workflow, the senior adjudicator resolved disagreements, provided clarification on ambiguous tagging scenarios, and ensured alignment with US-GAAP definitions and XBRL best practices.

Together, this composition provides both annotation robustness and domain oversight: junior analysts contribute detailed review capacity, while the senior adjudicator ensures conceptual correctness and methodological consistency across the audit.

All annotators were members of the research team and were invited to participate as part of their normal research training and collaborative project responsibilities. Participation was voluntary, and no separate monetary compensation was provided. Because the work involved publicly available financial disclosures and did not require annotators to process personal or sensitive data, the review posed minimal risk to participants.

\subsection{Validation Guideline}
\label{val_guide}

\subsubsection{Task Definition}

Given a context and an extracted triplet consisting of:
\begin{itemize}
    \item an \textbf{entity},
    \item its \textbf{entity type},
    \item a \textbf{US-GAAP concept},
\end{itemize}
Your task is to determine whether the triplet is correct.

Make a binary decision:
\begin{itemize}
    \item 1 for correct,
    \item 0 for incorrect.
\end{itemize}

\subsubsection{Validation Procedure}
Follow the rules below, the definitions of entity type are shown in Table~\ref{tab:entity-type-def}:

\begin{enumerate}
    \item \textbf{Entity Check}  
    Determine whether the extracted entity is a numerical value.  
    \begin{itemize}
        \item If yes, proceed to Step~2.  
        \item If no, set \texttt{is\_correct = 0}.
    \end{itemize}

    \item \textbf{Entity Type Validation}  
    Verify whether the identified entity type is correct based on the context and the definitions in Table~\ref{tab:entity-type-def}.  
    \begin{itemize}
        \item If yes, proceed to Step~3.  
        \item If no, set \texttt{is\_correct = 0}.
    \end{itemize}

    \item \textbf{US-GAAP Concept Validation}  
    Assess whether the assigned US-GAAP concept is appropriate based on the taxonomy tag definitions.  
    \begin{itemize}
        \item If yes, set \texttt{is\_correct = 1}.  
        \item If no, set \texttt{is\_correct = 0}.
    \end{itemize}
\end{enumerate}

\begin{table}[h]
\centering
\caption{Entity type definitions used for validation.}
\label{tab:entity-type-def}
\resizebox{\linewidth}{!}{
\begin{tabular}{lp{10cm}}
\toprule
\textbf{Entity Type} & \textbf{Definition} \\ \midrule
\texttt{monetaryItemType} & Financial amounts expressed in currency, such as revenue, profit, or total assets. \\
\texttt{integerItemType} & Counts of discrete items, such as the number of employees or total transactions. \\
\texttt{perShareItemType} & Per-share values, such as earnings per share (EPS) or book value per share. \\
\texttt{sharesItemType} & Counts of shares, such as outstanding shares or ownership stakes. \\
\texttt{percentItemType} & Ratios or percentages, such as tax rates, growth rates, or discount rates, usually expressed with a percentage symbol (\%). \\
\bottomrule
\end{tabular}
}
\end{table}

\subsubsection{Important Instructions}
\begin{enumerate}
    \item \textbf{Non-Arabic Formats}  
    Financial numerical entities may appear in word form (e.g., \textit{ten million}) and must be correctly identified and converted into standard numerical format.
    
    \item \textbf{Magnitude Terms}  
    If a number is followed by a magnitude term (e.g., \textit{hundred}, \textit{million}, \textit{billion}), do not expand it into the full numerical value:
    \begin{itemize}
        \item \textit{Two hundred} $\rightarrow$ extract only \textit{two}, not \textit{200}.  
        \item \textit{10.6 million} $\rightarrow$ extract only \textit{10.6}, not \textit{10,600,000}.
    \end{itemize}

    \item \textbf{Standardization of Format}  
    Remove formatting symbols while preserving the numerical value:
    \begin{itemize}
        \item Remove currency symbols (e.g., \textit{USD}).  
        \item Remove percentage signs (e.g., \%).  
        \item Remove commas (e.g., \textit{1,000} $\rightarrow$ \textit{1000}).
    \end{itemize}
\end{enumerate}

\subsection{Agreement Computation}
\label{agg_score}
We report both raw agreement and Cohen's $\kappa$ to quantify inter-annotator reliability.

Let $N$ denote the total number of instances, and let $n_{yy}$ be the number of cases
on which both annotators assign the same label $y$. The raw agreement is

\begin{equation}
P_o = \frac{1}{N} \sum_{y} n_{yy}.
\end{equation}

To adjust for agreement occurring by chance, we compute Cohen’s $\kappa$.
Let $p_{1y}$ and $p_{2y}$ denote the marginal probabilities that annotator~1 and annotator~2
assign label $y$, respectively. The expected chance agreement is

\begin{equation}
P_e = \sum_{y} p_{1y} \, p_{2y},
\end{equation}

and Cohen’s $\kappa$ is

\begin{equation}
\kappa = \frac{P_o - P_e}{1 - P_e}.
\end{equation}

Following \citet{landis1977measurement}, our $\kappa = 0.81$ indicates substantial agreement.

\section{Evaluation Metrics}
\label{metrics}

To provide a fair evaluation of overall benchmark performance, we adopt a set of metrics, focusing primarily on macro-level and micro-level evaluation strategies inspired by the previous work~\cite{sharma2023financial}.
\textbf{Macro-level} evaluation computes precision, recall, and F1 scores independently for each BIO-concept label derived from the US-GAAP taxonomy, and then averages them without weighting. This ensures that each concept, including rare or infrequent ones, contributes equally to the final score, making it especially suitable for domains with skewed label distributions. In contrast, \textbf{micro-level} evaluation aggregates token-level true positives, false positives, and false negatives across all labels before computing precision, recall, and F1. This approach emphasizes the model’s overall tagging accuracy by treating every token equally and thus better reflects performance on frequent concepts. Together, these two metrics provide a balanced view of both per-concept performance and overall tagging quality.

For the FinNI subtask, the objective is to extract correct \textbf{(entity, type)} pairs, that is $(\texttt{Fact}, \texttt{Type})$, from the financial document. Let $\mathcal{G} = {(e_i, l_i)}$ denote the set of ground-truth (entity, type) pairs, and $\mathcal{P} = {(e'_i, l'_i)}$ denote the set of predicted (entity, type) pairs. We evaluate the performance based on the following metrics:

\begin{equation} \text{Precision} = \frac{|\mathcal{G} \cap \mathcal{P}|}{|\mathcal{P}|} \end{equation}

\begin{equation} \text{Recall} = \frac{|\mathcal{G} \cap \mathcal{P}|}{|\mathcal{G}|} \end{equation}

\begin{equation} \text{F1} = \frac{2 \times \text{Precision} \times \text{Recall}}{\text{Precision} + \text{Recall}} \end{equation}
where $(e_i, l_i) = (e'_j, l'_j)$ if and only if both the entity span $e$ and its assigned type $l$ exactly match.

For the FinCL subtask, Given a set of queries $\mathcal{Q} = \{q_1, q_2, \dots, q_N\}$, where each $q_i$ is associated with a ground-truth concept $c_i^*$ and a predicted concept $\hat{c}_i$, the accuracy is defined as:

\begin{equation} \text{Accuracy} = \frac{1}{N} \sum_{i=1}^{N} \delta(\hat{c}_i, c_i^*) \end{equation}
where $\delta(\hat{c}_i, c_i^*) = 1$ if $\hat{c}_i = c_i^*$, and $0$ otherwise.

\section{Evaluation Models Details}
\label{model_list}

Table~\ref{model-table} provides an overview of the models evaluated in this study, categorized by openness, domain specialization, and architectural foundation. The evaluation covers a diverse range of models:

\begin{itemize}
    \item Closed-source LLMs: We include GPT-4o~\cite{hurst2024gpt}, accessed via OpenAI’s API, as a representative of cutting-edge proprietary models with demonstrated performance across a variety of NLP tasks. Although model size details are undisclosed, GPT-4o serves as an upper-bound reference in our benchmark.
    \item Open-source LLMs: This group encompasses recent, high-performing open models such as DeepSeek-V3 (685B)~\cite{liu2024deepseek}, DeepSeek-R1-Distill-Qwen (32B)~\cite{deepseekai2025deepseekr1incentivizingreasoningcapability}, and multiple variants of Qwen3 ~\cite{qwen3technicalreport}(ranging from 0.6B to 32B). We also include Llama-3.3, Llama-3.2, and 3.1 variants ~\cite{grattafiori2024llama} (70B, 3B, and 8B), Llama-4-Scout-17B-16E-Instruct~\cite{meta2025llama}(109B), as well as Google's gemma-2-27B ~\cite{gemma_2024}, to ensure architectural diversity and scalability comparison. These models are primarily instruction-tuned and optimized for general-purpose NLP tasks.
    \item Financial-specific LLMs: We evaluate Fino1-8B ~\cite{qian2025fino1}, a domain-specialized model trained on financial corpora, designed to better capture the terminology and structure unique to financial disclosures. This category allows us to assess the benefits of domain adaptation in complex tagging and reasoning tasks.
    \item Pretrained Language Models (PLMs): To establish strong baselines, we include non-generative encoder models: BERT-large ~\cite{DBLP:journals/corr/abs-1810-04805}, FinBERT ~\cite{araci2019finbert}, and SECBERT ~\cite{loukas2022finer}. These models have been widely used in prior financial NLP tasks and allow for a comparative analysis between generative LLMs and traditional pretrained models in terms of domain understanding and structured output capability.
\end{itemize}

Together, these models offer a comprehensive evaluation spectrum, from general-purpose to domain-specific, encoder-based to decoder-based, and open to closed source, facilitating an in-depth assessment of their performance across our proposed benchmark tasks.

\begin{table}[!h]
\scriptsize
\caption{Model categories and corresponding repositories.}
\label{model-table}
\centering
\resizebox{\linewidth}{!}{%
\begin{tabular}{lll}
    \toprule
    Model & Size & Source \\
    \midrule
    \textbf{Closed-source Large Language Models} \\
    GPT-4o & - & gpt-4o-2024-08-06 \\
    \midrule
    \textbf{Open-source Large Language Models} \\
    DeepSeek-V3 & 685B & deepseek-ai/DeepSeek-V3 \\
    Llama-4-Scout-17B-16E-Instruct & 109B & meta-llama/Llama-4-Scout-17B-16E-Instruct \\
    Llama-3.3-70B-Instruct & 70B & meta-llama/Llama-3.3-70B-Instruct \\
    DeepSeek-R1-Distill-Qwen & 32B & deepseek-ai/DeepSeek-R1-Distill-Qwen-32B \\
    Qwen3-32B & 32B & Qwen/Qwen3-32B \\
    gemma-2-27b-it & 27B & google/gemma-2-27b-it \\
    Qwen3-14B & 14B & Qwen/Qwen3-14B \\
    Llama-3.1-8B-Instruct & 8B & meta-llama/Llama-3.1-8B-Instruct \\
    Llama-3.2-3B-Instruct & 3B & meta-llama/Llama-3.2-3B-Instruct \\
    Qwen3-1.7B & 1.7B & Qwen/Qwen3-1.7B \\
    Qwen3-0.6B & 0.6B & Qwen/Qwen3-0.6B \\
    \midrule
    \textbf{Financial-specific Large Language Models} \\
    Fino1 & 8B & TheFinAI/Fino1-8B \\
    \midrule
    \textbf{Pretrained Language Models} \\
    BERT-large & $\sim$340M & google-bert/bert-large-uncased \\
    FinBERT & $\sim$110M & ProsusAI/finbert \\
    SECBERT & $\sim$110M & nlpaueb/sec-bert-base \\
    \bottomrule
\end{tabular}%
}
\end{table}

\section{The details for the fine-tuning PTMs}
\label{training_set}
\subsection{Training data collection and processing}
Similar to the collection process for the \textsc{\fintagging} benchmark data, we gathered an additional 10 annual 10-K financial reports filed with the SEC for the period from February 13, 2024, to February 13, 2025, as summarized in Table~\ref{statistic_traindata}. These reports contain a total of 33,848 standard taxonomy-tagged facts. Using BeautifulSoup to parse these documents, we identified 22,847 narrative sentences (approximately 5.5 million characters) and 1,236 financial tables. The companies included in this dataset follow the XBRL standard, ensuring comprehensive coverage for training PTMs.

\begin{table}[htbp]
\caption{Financial report statistics summary for raw training data}
\label{statistic_traindata}
\centering
\scriptsize
\begin{tabular}{ll}
\toprule
\textbf{Item} & \textbf{Information} \\
\midrule
Report type     & 10-K \\
Period          & 2024-02-13 to 2025-02-13 \\
\#Company       & 10 \\
\#Sentence      & 22,847 \\
\#Table         & 1,236 \\
\#Characters    & 5,539,198 \\
\#Standard Tags & 33,848 \\
\bottomrule
\end{tabular}
\end{table}

After collection, we employed the same procedure to filter texts and tables, subsequently annotating numerical entities, entity types, and US-GAAP tags (concepts). Finally, as detailed in Table~\ref{tab:trainset_stats}, we generated a total of 1,116 sentences and 953 tables as the training set for PTMs. Specifically, the sentence-level data consists of 558 positive and 558 negative instances, averaging approximately 84.24 tokens ($\pm$ 69.29), with 1.22 annotated entities and concepts per sentence. The table-level data comprises 594 positive and 359 negative instances, with a significantly higher average of 1,281.86 tokens ($\pm$ 6,438.37), and approximately 25 entities and concepts annotated per table. Overall, the annotated dataset includes 25,199 entities, covering 1,435 unique US-GAAP concepts.

\begin{table}[t]
  \caption{Statistics of training data (tokens calculated with ``cl100k\_base'' tokenizer, $\pm$ standard deviation).}
  \label{tab:trainset_stats}
  \centering
  \resizebox{\columnwidth}{!}{%
  \begin{tabular}{llcccccc}
    \toprule
    Structure & Pos/Neg & \#Instance & Avg. Tokens/S & Avg. Entities/S & Avg. Concepts/S & Total Entities & Unique Concepts \\
    \midrule
    \multirow{2}{*}{Sentence}  
             & Positive & 558 & \multirow{2}{*}{$84.24 \pm 69.29$} & \multirow{2}{*}{$1.22 \pm 1.78$} & \multirow{2}{*}{$1.22 \pm 1.78$} & \multirow{4}{*}{25,199} & \multirow{4}{*}{1,435} \\
             & Negative & 558 &  &  &  &  &  \\
    \multirow{2}{*}{Table} 
             & Positive & 594 & \multirow{2}{*}{$1281.86 \pm 6438.37$} & \multirow{2}{*}{$25.00 \pm 213.77$} & \multirow{2}{*}{$25.00 \pm 213.77$} &  &  \\
             & Negative & 359 &  &  &  &  &  \\
    \bottomrule
  \end{tabular}%
  }
\end{table}

\begin{figure*}[htbp]
  \centering
  \includegraphics[width=.8\linewidth]{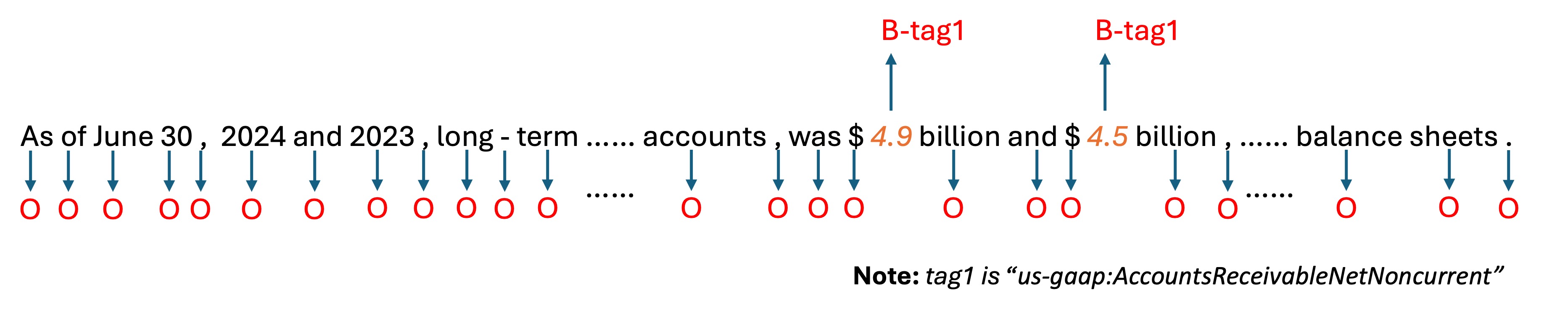}
  \caption{An example for training set annotation with BIO scheme.}
  \label{fig:bio}
\end{figure*}

However, to align with the extreme classification format used in previous XBRL tagging benchmarks, we directly adopt the US-GAAP tags as entity labels, annotating each token in sentences and tables using the \texttt{BIO} scheme. Specifically, \texttt{B} denotes the beginning of an entity phrase, \texttt{I} marks the continuation (inside) of an entity phrase, and \texttt{O} indicates tokens outside of any entity. As shown in Figure~\ref{fig:bio}, "4.9" and "4.5" are single-token numerical entities labeled only with a \texttt{B} prefix (e.g., ``B-us-gaap:AccountsReceivableNetNoncurrent''). To comprehensively cover all US-GAAP tags, we combine the entire set of 17,688 tags from the US-GAAP 2023 and 2024 taxonomy with the \texttt{BIO} labeling scheme to construct an extreme classification label space, resulting in 34,777 unique entity labels ($2 \times 17388 + 1$).

After constructing the training set, we reconstruct the testing set from the original benchmark dataset. The training settings are detailed below.

\subsection{Training settings}
\label{tran_setting}

We fine-tune three pretrained models, BERT-large~\cite{DBLP:journals/corr/abs-1810-04805}, FinBERT~\cite{araci2019finbert}, and SECBERT~\cite{loukas2022finer}, on our training set using the HuggingFace Transformers library. All models are trained with a batch size of 4, a learning rate of 3e-5, and for 20 epochs. Optimization is performed using AdamW without gradient accumulation or early stopping. Token classification heads are randomly initialized and trained jointly with the base encoder. Input sequences are tokenized with a maximum length of 512, and labels are aligned at the sub-token level following the BIO tagging scheme. Loss is computed only on the first sub-token of each word to avoid misalignment bias.

Training is conducted on two NVIDIA A5000 GPUs (24GB each) using data parallelism for 24 hours. All other hyperparameters follow the default settings in the HuggingFace Trainer API. Models are evaluated using the checkpoint from the final epoch. All experiments are run under a fixed random seed to ensure reproducibility.

\section{Retrieval Results in FinCL subtask}
\label{retrieval_res}
We investigate the impact of different context construction strategies for the queried entity at the \textbf{retrieval stage}. We consider two approaches: Fixed-Window Context (FWC) and Structure-Aware Context (SAC). In the FWC strategy, context is constructed by extracting a fixed window of 50 characters before and after the entity mention, regardless of whether the entity appears in a sentence or a table. In contrast, the SAC strategy builds context based on the structural location of the entity: if the entity appears in a sentence, the entire sentence is used; if the entity appears in a table, we linearize the entire row into a Markdown-style key-value format to serve as the context.

\begin{table}[t]
  \caption{Acc@k retrieval performance on the FinCL task. FWC: fixed window-based; SAC: structure-aware.}
  \label{tab:retrieval_res}
  \centering
  \resizebox{\columnwidth}{!}{%
  \begin{tabular}{lcccccccc}
    \toprule
    Strategy & Structure & Acc@1 & Acc@10 & Acc@20 & Acc@50 & Acc@100 & Acc@150 & Acc@200  \\
    \midrule
    \multirow{3}{*}{FWC} 
    & Sentence & 0.0658 & 0.2614 & 0.3562 & 0.4650 & 0.5618 & 0.6114 & 0.6502 \\
    & Table    & 0.0000 & 0.0029 & 0.0038 & 0.0042 & 0.0050 & 0.0065 & 0.0089 \\
    & Overall  & 0.0055 & 0.0237 & 0.0316 & 0.0409 & 0.0492 & 0.0544 & 0.0608 \\
    \midrule
    \multirow{3}{*}{SAC} 
    & Sentence & 0.0696 & 0.2742 & 0.3631 & 0.4798 & 0.5701 & 0.6125 & 0.6535 \\
    & Table    & 0.0159 & 0.0872 & 0.1245 & 0.1938 & 0.2452 & 0.2723 & 0.2959 \\
    & Overall  & 0.0202 & 0.1152 & 0.1534 & 0.2188 & 0.2727 & 0.3012 & 0.3274 \\
    \bottomrule
  \end{tabular}%
  }
\end{table}

From Table~\ref{tab:retrieval_res}, we observe that the SAC strategy consistently outperforms FWC, particularly in the table context, highlighting the importance of aligning the context window with the underlying structural unit.

For sentence-based entities, both strategies yield comparable results, with SAC achieving slightly higher Acc@1 (0.0696 vs. 0.0658) and showing marginal improvements across all cutoff values (e.g., Acc@100 of 0.5701 vs. 0.5618). This suggests that even for relatively unstructured text, preserving sentence boundaries provides minor benefits over fixed-length windows.

In contrast, for table-based entities, the performance gap is substantial. SAC achieves significantly higher retrieval accuracy (e.g., Acc@100 of 0.2452 vs. 0.0050), indicating that row-level context is far more informative than arbitrary character windows when dealing with tabular structures. FWC performs poorly in this setting, likely due to the fragmented and semantically sparse nature of partial table text.

When aggregating results across both structures, SAC outperforms FWC by a wide margin at all retrieval depths (e.g., Acc@200 of 0.3274 vs. 0.0608). These findings underscore the importance of structure-aware context construction, especially in scenarios where inputs span multiple formats such as sentences and tables.

\section{Error cases}
\label{error_cases}
To better illustrate the semantic ambiguity challenge, we present a representative error case from our best-performing model, DeepSeek-V3. Consider the following input paragraph: 

{\textit{``Cash equivalents include term deposits with banks, money market funds, and all highly liquid investments with original maturities of three months or less...... At December 28, 2024, we had restricted cash of \textbf{\$31 million} recorded in other current assets and restricted cash of \textbf{\$121 million} recorded in other non-current assets.''}

This paragraph contains 2 ``monetary entities'': 31 and 121. The gold concepts and predicted concepts that are assigned to both entities are shown in Table~\ref{tab:error_cases}. For case 1, the model’s choice was likely influenced by the phrase ``recorded in current assets''. Yet the paragraph describes the overall composition of restricted cash, and the correct concept reflects this aggregate view rather than a single classification. For case 2, although both concepts denote non-current restricted cash, the model selected a narrower variant, showing the difficulty of distinguishing between highly similar concepts with subtle differences in scope.

\begin{table}[t]
\setlength{\abovecaptionskip}{3pt}  
\centering
\caption{Error cases from DeepSeek-V3 rerank.}
\label{tab:error_cases}
\scriptsize
\begin{tabularx}{\linewidth}{lX}
\toprule
\textbf{Case 1: \$31m} &  \\
\midrule
Gold Concept & \text{us-gaap:RestrictedCashAndCashEquivalentsAtCarryingValue} \\
Gold Concept Rank & 18 / 200 (not in Top-5) \\
Model Prediction & \text{us-gaap:RestrictedCashAndCashEquivalentsCurrent} \\
\midrule
\textbf{Case 2: \$121m} &  \\
\midrule
Gold Concept & \text{us-gaap:RestrictedCashAndCashEquivalentsNoncurrent} \\
Gold Concept Rank & 2 / 200 (in Top-5) \\
Model Prediction & \text{us-gaap:CashCashEquivalentsRestrictedCashAndRestrictedCashEquivalents} \\
\bottomrule
\end{tabularx}
\end{table}

\newpage

\begin{figure*}[t]
\centering
\begin{tcolorbox}[colback=lightgray!10, colframe=black, title=Prompt Template for FinNI Subtask, width=\textwidth]
\fontsize{8pt}{9.8pt}
\begin{lstlisting}[breaklines=true, basicstyle=\ttfamily, frame=none]
You are a financial information extraction expert specializing in identifying financial numerical entities in XBRL reports.
Your task is to extract all such numerical entities from the provided text or serialized <table></table> data and classify them into one of five categories:
        
        - "integerItemType": Counts of discrete items, such as the number of employees or total transactions.
        - "monetaryItemType": Financial amounts expressed in currency, such as revenue, profit, or total assets.
        - "perShareItemType": Per-share values, such as earnings per share (EPS) or book value per share.
        - "sharesItemType": Counts of shares, such as outstanding shares or ownership stakes.
        - "percentItemType": Ratios or percentages, such as tax rates, growth rates, or discount rates, usually expressed with a percentage symbol ("%").
    
Important Instructions:
        (1) Financial numerical entities are not limited to Arabic numerals (e.g., 10,000). They may also appear in word form (e.g., "ten million"), which must be correctly identified and converted into standard numerical format.
        (2) Not all numbers in the text should be extracted. Only those that belong to one of the five financial entity categories above should be included. Irrelevant numbers (such as phone numbers, dates, or general IDs) must be ignored.
        (3) If a number is followed by a magnitude term (e.g., Hundred, Thousand, Million, Billion), do not expand it into the full numerical value.
            * "Two hundred" -> Extract only "two", not "200".
            * "10.6 million" -> Extract only "10.6", not "10,600,000".
        (4) Standardize numerical formatting by removing currency symbols (e.g., "USD"), percentage signs ("%"), and commas (",") while preserving the numeric value. These elements must be removed to ensure consistency.
        (5) Output the extracted financial entities in JSON list format without explanations, structured as follows: {"result":[{"Fact": <Extracted Numerical Entity>, "Type": <Identified Entity Type>}]}
    
Input: {text/table}
Output:
\end{lstlisting}
\end{tcolorbox}
\caption{Prompt template used for the FinNI subtask.}\label{tem-finni}
\end{figure*}

\clearpage
\begin{figure*}[t]
\centering
\begin{tcolorbox}[colback=lightgray!10, colframe=black, title=Prompt Template for FinCL Subtask (Reranking), width=\textwidth]
\fontsize{8pt}{9.8pt}
\begin{lstlisting}[breaklines=true, basicstyle=\ttfamily, frame=none]
You are a financial tagging assistant trained in US-GAAP taxonomy.

Given a query consisting of an entity, its type, its surrounding context, and the source format (either text or table), your task is to select the single most appropriate US-GAAP tag from a list of 200 candidate tags.

Make your decision by carefully analyzing the meaning and context of the entity and matching it with the semantics of the tags.

Only output one tag, the best match. Do not explain or list multiple tags. The output is a JSON format: {"result": <the best matched tag>}.

Input Query: <entity> + <entity type> + <context>
Candidate Tags: {Top 200 US-GAAP tags}

Answer:
\end{lstlisting}
\end{tcolorbox}
\caption{Prompt template used for the Reranking stage in the FinCL subtask.}\label{tem-fincl}
\end{figure*}

\clearpage

\begin{figure*}[t]
\centering
\begin{tcolorbox}[colback=lightgray!10, colframe=black, title=Prompt Template for Ablation, width=\textwidth]
\fontsize{8pt}{9.8pt}
\begin{lstlisting}[breaklines=true, basicstyle=\ttfamily, frame=none]
You are an XBRL tagging expert specializing in annotating financial numerical facts in XBRL reports.
Your task is to (1) extract all such numerical entities from the provided text or serialized <table></table> data, (2)  classify them into one of five categories, and (3) assign an appropriate US-GAAP tag to each entity.

Categories:   
        - "integerItemType": Counts of discrete items, such as the number of employees or total transactions.
        - "monetaryItemType": Financial amounts expressed in currency, such as revenue, profit, or total assets.
        - "perShareItemType": Per-share values, such as earnings per share (EPS) or book value per share.
        - "sharesItemType": Counts of shares, such as outstanding shares or ownership stakes.
        - "percentItemType": Ratios or percentages, such as tax rates, growth rates, or discount rates, usually expressed with a percentage symbol ("%").
    
US-GAAP tags:
       - A US-GAAP tag is a standardized semantic label used in XBRL filings to identify specific financial concepts defined by the U.S. Generally Accepted Accounting Principles (GAAP). Each tag represents a distinct accounting item and enables consistent, machine-readable financial reporting.
       - Examples: "us-gaap:AssetsCurrentAbstract", "us-gaap:AccruedInsuranceNoncurrent".

Important Instructions:
        (1) Financial numerical entities are not limited to Arabic numerals (e.g., 10,000). They may also appear in word form (e.g., "ten million"), which must be correctly identified and converted into standard numerical format.
        (2) Not all numbers in the text should be extracted. Only those that belong to one of the five financial entity categories above should be included. Irrelevant numbers (such as phone numbers, dates, or general IDs) must be ignored.
        (3) If a number is followed by a magnitude term (e.g., Hundred, Thousand, Million, Billion), do not expand it into the full numerical value.
            * "Two hundred" -> Extract only "two", not "200".
            * "10.6 million" -> Extract only "10.6", not "10,600,000".
        (4) Standardize numerical formatting by removing currency symbols (e.g., "USD"), percentage signs ("%"), and commas (",") while preserving the numeric value. These elements must be removed to ensure consistency.
        (5) You should assign the most appropriate US-GAAP tag to each identified entity based on your internal understanding of the 2024 US-GAAP taxonomy. 
        (6) Output the extracted financial entities in JSON list format without explanations, structured as follows: {"result":[{"Fact": <Extracted Numerical Entity>, "Type": <Identified Entity Type>, "Tag": <Assigned US-GAAP tag>}]}
    
Input: {text/table}
Output:

\end{lstlisting}
\end{tcolorbox}
\caption{Prompt template used for ablation study.}
\label{tem-abl}
\end{figure*}

\end{document}